\UseRawInputEncoding
\documentclass[manuscript,screen]{acmart}
\AtBeginDocument{%
  }

\usepackage{colortbl}

\usepackage{amsmath,amssymb}
\usepackage{algorithmic}
\usepackage{textcomp}
\usepackage{xcolor}
\usepackage{colortbl}
\usepackage{booktabs}
\usepackage{multirow}
\usepackage{placeins}
\usepackage{subcaption}
\usepackage{alltt}
\usepackage[most]{tcolorbox}
\usepackage{tcolorbox} 
\usepackage{fvextra}
\tcbuselibrary{breakable} 
\usepackage{xcolor}
\usepackage{fancyvrb}
\usepackage{graphicx}
\usepackage{enumitem}
\usepackage{geometry}
\usepackage{listings} 

\begin{document}

\title{ReGraph: Learning to Generate Recipe Graphs from Food Images}

\author{Guoshan Liu}
\affiliation{%
  \institution{Institute of Trustworthy Embodied AI, Fudan University}
  \city{Shanghai}
  \country{China}
}
\affiliation{%
  \institution{Shanghai Key Laboratory of Multimodal Embodied AI}
  \city{Shanghai}
  \country{China}
}
\email{gsliu24@m.fudan.edu.cn}

\author{Bin Zhu}
\affiliation{%
  \institution{Singapore Management University}
  \city{Singapore}
  \country{Singapore}
}
\email{binzhu@smu.edu.sg}

\author{Pengkun Jiao}
\affiliation{%
  \institution{Institute of Trustworthy Embodied AI, Fudan University}
  \city{Shanghai}
  \country{China}
}
\affiliation{%
  \institution{Shanghai Key Laboratory of Multimodal Embodied AI}
  \city{Shanghai}
  \country{China}
}
\email{pkjiao23@m.fudan.edu.cn}

\author{Jingjing Chen}
\authornote{Corresponding author.}
\affiliation{%
  \institution{Institute of Trustworthy Embodied AI, Fudan University}
  \city{Shanghai}
  \country{China}
}
\affiliation{%
  \institution{Shanghai Key Laboratory of Multimodal Embodied AI}
  \city{Shanghai}
  \country{China}
}
\email{chenjingjing@fudan.edu.cn}

\author{Chong-Wah Ngo}
\affiliation{%
  \institution{Singapore Management University}
  \city{Singapore}
  \country{Singapore}
}
\email{cwngo@smu.edu.sg}

\author{Yu-Gang Jiang}
\affiliation{%
  \institution{Institute of Trustworthy Embodied AI, Fudan University}
  \city{Shanghai}
  \country{China}
}
\affiliation{%
  \institution{Shanghai Key Laboratory of Multimodal Embodied AI}
  \city{Shanghai}
  \country{China}
}
\email{ygj@fudan.edu.cn}

\renewcommand{\shortauthors}{Liu et al.}

\begin{abstract}
  Recent Large Multimodal Models (LMMs) have achieved impressive performance in recipe generation from food images. However, cooking is a structured transformation process in which ingredients undergo state changes through ordered actions, while free-form recipe language leaves the corresponding entities, intermediate states, and dependencies largely implicit and entangled. A graph representation makes this procedural knowledge explicit and compositional, providing a structured basis for assessing whether model outputs encode process-level knowledge rather than merely presenting plausible textual descriptions. To address this limitation, we present ReGraph, a large-scale recipe graph dataset that represents ingredients, cooking actions, and tools as entities, uses entity attributes to describe ingredient state changes, and employs typed relations to encode manipulation targets, destinations, and procedural ordering. ReGraph further incorporates explicit Recipe Reasoning Chain-of-Thought (RR-CoT) traces, providing auxiliary supervision for procedural decomposition and structured graph generation. Building on ReGraph, we propose Recipe Graph Learning (RGL), a two-stage framework that enables LMMs to generate a plausible fine-grained cooking workflow from a food image in the form of a structured recipe graph. Under a deterministic, schema-aware matching protocol, our experiments reveal a substantial gap between text-generation quality and recoverable procedural structure: recipes produced by existing approaches achieve competitive text-generation scores yet yield limited reference-aligned entity and relation structure under the ReGraph schema. In contrast, across two representative LMM backbones, RGL consistently improves the generation of cooking entities and procedural relations, while our analysis further shows that fine-grained ingredient-state capture remains the most challenging dimension. Together, ReGraph and RGL provide a new dataset, benchmark, and learning framework for explicit recipe process modeling and structured procedural understanding.
\end{abstract}


\maketitle

\section{Introduction}
\label{intro}
\begin{figure}[t]
\centering
\includegraphics[width=0.7\columnwidth]{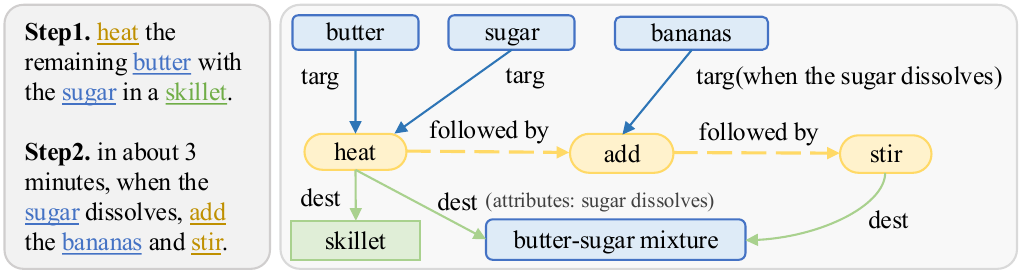} 
\caption{A sample visualization showing original recipe instructions alongside a corresponding partial entry from ReGraph. Nodes represent ingredients (blue), actions (yellow), and tools (green). Relations define procedural interactions: \texttt{targ} links an ingredient or other directly manipulated entity to the action applied to it, \texttt{dest} links an action to its destination or output entity (e.g., a tool such as skillet or a newly formed ingredient entity such as butter-sugar mixture), and \texttt{followed by} captures temporal and procedural ordering between actions. Ingredient state transformations are represented through entity attributes, while the graph jointly captures ingredient evolution and step-level procedural structure derived from the recipe instructions.}
\label{kg_case}
\vspace{-0.8em}
\end{figure}

The growing interest in healthy living has catalyzed significant advancements in food computing \cite{foodsurvey, chen2020study, song2024enhancing, gui2024navigating, wu2026dual, zhang2025cookanything, qi2025advancing}, particularly in ingredient parsing \cite{gyx, liu2024canteen}, cross-modal recipe retrieval \cite{chen2016deep, wahed2024fine}, and recipe generation \cite{RecipeGPT, liu2025retrieval, liu2026enhancing}. The recent paradigm shift toward Large Multimodal Models (LMMs) has further accelerated this progress. By grounding broad culinary knowledge in visual semantics, LMMs can now generate detailed and plausible recipes from single food images. Representative examples include FoodLMM \cite{foodlmm}, which focuses on fine-grained food identification and understanding, and RecipeRAG \cite{reciperag}, which employs retrieval-augmented generation to improve the consistency of generated recipes. 

Existing methods have achieved competitive performance on conventional recipe generation benchmarks, where generation quality is commonly assessed using text-based metrics such as SacreBLEU \cite{sacrebleu} and ROUGE-L \cite{lin2004rouge}. However, fluent and lexically similar recipe text does not necessarily provide explicit representation of fine-grained cooking procedures. Fine-grained information such as how ingredient states evolve after individual actions, which ingredients or intermediate products are manipulated by each action, and which actions must precede others is often embedded implicitly in free-form instructions. Consequently, generated recipes may be readable and semantically plausible while remaining difficult to analyze, compare, or reason over at the procedural level.

Structured graphs provide a natural knowledge representation for cooking procedures. Recipes describe sequences of ingredient-state transitions induced by cooking actions, but free-form instructions distribute this knowledge across narrative expressions, leaving the participating entities, intermediate states, and procedural dependencies implicit. By encoding these elements as nodes, attributes, and typed relations, a recipe graph makes the transformation process explicit and compositional. It therefore supports direct process-level analysis and provides a structured proxy for assessing whether generated outputs encode procedural knowledge beyond textual plausibility \cite{graphsurvey, khoshraftar2024survey}. Nevertheless, as summarized in Table~\ref{tab:dataset_comparison}, existing food-domain graph resources \cite{foodkg, 2014_kg, 2020_kg, 2022_kg} remain limited in scale, procedural completeness, or explicit procedural reasoning supervision. Some primarily encode static ingredient, nutrition, or cross-recipe associations, whereas others contain only partial cooking flows or represent intermediate states implicitly. These limitations make them unsuitable for large-scale training and evaluation of LMMs that generate fine-grained structured cooking workflows from visual inputs.
\begin{table*}[htbp]
\centering
\setlength{\abovecaptionskip}{1pt}
\caption{Comparison between ReGraph and existing food graph datasets. We emphasize the advantages of ReGraph in modeling fine-grained procedural dependencies and explicit ingredient state transformations.}
\label{tab:dataset_comparison}
\resizebox{\textwidth}{!}{
\begin{tabular}{lccccc}
\toprule
\textbf{Dataset} & \textbf{Recipes} & \textbf{Relations / Edges} & \textbf{Step Dependency} & \textbf{Ingredient State Transformation} \\ 
\midrule
Flow Graph Corpus \cite{2014_kg} & 266 & -- & $\checkmark$ & Limited \\
Visual Grounding Annotation \cite{2020_kg} & 272 & 2,300 (region annotations) & $\times$ & Limited \\
Visual Recipe Flow \cite{2022_kg} & 200 & 11,291 (edges) & $\checkmark$ & Partial \\
FoodKG \cite{foodkg} & >1,000,000 & >67,000,000 & $\times$ & $\times$ \\
Cooking Programs \cite{papadopoulos2022learning} & 3,708 & 54,154 (edges) & $\checkmark$ & Latent \\
\midrule
\rowcolor{gray!10} \textbf{ReGraph (Ours)} & \textbf{10,000} & \textbf{391,051} & \textbf{$\checkmark$} & \textbf{Fine-grained} \\ 
\bottomrule
\end{tabular}
}
\vspace{-0.8em}
\end{table*}

To address these limitations, we introduce ReGraph, a large-scale recipe graph dataset containing 10,000 recipes sampled from Recipe1M \cite{recipe1m}, with 318,773 entities and 391,051 relations. Each sample contains two complementary components: (1) a Recipe Reasoning Chain-of-Thought (RR-CoT) trace that decomposes the recipe instructions into procedural steps and identifies their ordering relationships, and (2) a structured recipe graph composed of ingredient, action, and tool entities connected by the typed relations \texttt{targ}, \texttt{dest}, and \texttt{followed by}. Ingredient state changes and intermediate products are explicitly represented through newly instantiated entities and their attributes. The RR-CoT serves as auxiliary procedural decomposition supervision during supervised fine-tuning, whereas the recipe graph serves as the canonical knowledge representation and the primary prediction and evaluation target. Figure~\ref{kg_case} presents a representative partial example from ReGraph. By making fine-grained procedural information explicit, ReGraph addresses the representation-level limitations of free-form recipe text. Generating such structured information from food images, however, remains challenging because ingredient transformations, action--ingredient interactions, intermediate products, and procedural ordering cannot be directly observed from visual appearance alone.

Building on ReGraph, we propose Recipe Graph Learning (RGL), a two-stage framework for generating structured cooking workflows from food images. In the first stage, RGL-SFT jointly supervises the generation of an RR-CoT trace and its corresponding recipe graph, encouraging the model to organize the inferred cooking procedure before producing structured entities and relations. In the second stage, RGL-RFT applies Group Relative Policy Optimization (GRPO) \cite{GRPO} to further improve graph-level generation quality. Specifically, the proposed Relative Improvement Reward (RIR) promotes improvements in entity and relation prediction relative to the supervised baseline, while a lightweight Format Reward encourages parsable and schema-compliant outputs. To measure progress on this task, we adopt a deterministic, schema-aware canonical matching protocol that scores generated graphs against reference annotations without relying on model-based judgments for the primary metric. With this two-stage design, RGL consistently improves fine-grained recipe graph generation across two representative LMM backbones, Qwen3-VL-8B \cite{qwen3} and InternVL3-8B \cite{internvl}.
Our contributions can be summarized as follows:
\begin{itemize}[leftmargin=0.8em, itemsep=2pt, topsep=2pt]
    \item We introduce ReGraph, a large-scale recipe graph dataset that explicitly represents cooking entities, ingredient state evolution, intermediate products, and procedural ordering, together with RR-CoT traces for auxiliary reasoning supervision.
    \item We propose Recipe Graph Learning (RGL), a two-stage framework that enables LMMs to generate fine-grained structured cooking workflows directly from food images through supervised and graph-level reinforcement fine-tuning.
    \item We establish a systematic benchmark for recipe graph generation from food images and reveal a substantial gap between conventional text-generation quality and the amount of explicit procedural structure that can be systematically extracted and evaluated from generated recipes. Across two representative LMM backbones, RGL consistently improves graph-level entity and relation generation.
\end{itemize}

\section{Related Work}
\subsection{Recipe Generation with Large Multimodal Models}
Recent advances in Large Multimodal Models (LMMs) have substantially improved recipe generation from food images by leveraging stronger visual understanding and language generation capabilities. FoodLMM \cite{foodlmm} enhances fine-grained food recognition and recipe understanding through multimodal instruction tuning. RoDE \cite{rode} further improves food-domain multimodal learning through a unified food benchmark and a mixture-of-experts architecture that supports multiple food understanding tasks, including recipe generation. SDRA \cite{liu2025retrieval} incorporates semantic retrieval to provide more relevant external recipe knowledge for generation. Building on retrieval-based recipe modeling, RecipeRAG \cite{reciperag} further improves cross-modal recipe retrieval accuracy and adopts a two-stage training strategy combining supervised fine-tuning with GRPO-based reinforcement optimization \cite{GRPO} to improve conventional recipe generation performance. More recently, SGRG \cite{liu2026enhancing} argues that conventional text-generation metrics are insufficient for evaluating recipe generation quality and introduces semantic-level evaluation metrics to better assess the generated recipes. Nevertheless, existing approaches remain primarily optimized for textual quality and semantic consistency. Their free-form outputs do not explicitly expose fine-grained procedural information, such as action--ingredient interactions, tool usage, ingredient state evolution, intermediate products, and ordering constraints among cooking actions, thereby limiting direct process-level analysis and evaluation.
\subsection{Structured Generation with Large Multimodal Models}
Recent Large Multimodal Models (LMMs) \cite{qwen3,internvl,gpt52} have demonstrated remarkable capabilities in generating structured outputs from visual inputs beyond free-form natural language. Pix2Seq \cite{pix2seq} reformulates structured vision tasks as autoregressive sequence generation under a unified language modeling framework. More recently, Image2Struct \cite{image2struct} systematically benchmarks vision-language models on recovering structured source representations directly from images through image-to-structure generation. LMM-based approaches have further extended structured generation to machine-readable outputs such as executable programs and symbolic representations \cite{llm_relatedwork1,llm_relatedwork2}. 
Compared with open-ended text generation, structured output generation requires simultaneously preserving semantic correctness, structural consistency, and schema validity, making it considerably more challenging. Chain-of-Thought (CoT) prompting \cite{cot} has further improved structured prediction by decomposing complex reasoning into intermediate inference steps. Nevertheless, existing methods mainly focus on generating generic structured representations or executable programs for general visual reasoning. Extending structured output generation to procedural recipe graphs remains largely unexplored, as it requires inferring latent ingredient transformations, intermediate products, action--ingredient interactions, and procedural ordering that are not directly observable from the final food appearance alone.
\subsection{Graph Representations for Food Computing}
\label{sec:related_graph}
Structured graph representations \cite{graphsurvey,khoshraftar2024survey} provide a principled framework for modeling entities, relations, and procedural dependencies in complex reasoning tasks. Existing food-domain graph resources range from large-scale static knowledge graphs to smaller procedural representations. FoodKG \cite{foodkg}, a large-scale food-domain knowledge graph containing over one million recipes and 67 million triples, primarily models static ingredient properties, nutritional information, and cross-recipe associations without representing cooking workflows. To capture procedural information, \cite{2014_kg} introduced directed action graphs constructed from recipe text, although without visual grounding. Subsequently, \cite{2020_kg} incorporated ingredient bounding boxes for visual grounding, but did not explicitly model temporal dependencies or ingredient state transitions across cooking steps. \cite{2022_kg} further introduced state-change image pairs, providing richer supervision but focusing primarily on dataset construction rather than graph generation. More recently, \cite{papadopoulos2022learning} represented recipes as executable programs, where intermediate cooking states are treated as latent variables instead of explicit graph entities. As summarized in Table~\ref{tab:dataset_comparison}, existing food-domain graph resources remain limited in scale, procedural completeness, and explicit procedural reasoning supervision. ReGraph addresses these limitations by providing automatically constructed recipe graph annotations, with fully human-verified annotations for the evaluation split and semi-automatically refined annotations for the training split.
\begin{figure}[t]
    \centering
    \includegraphics[width=0.8\linewidth]{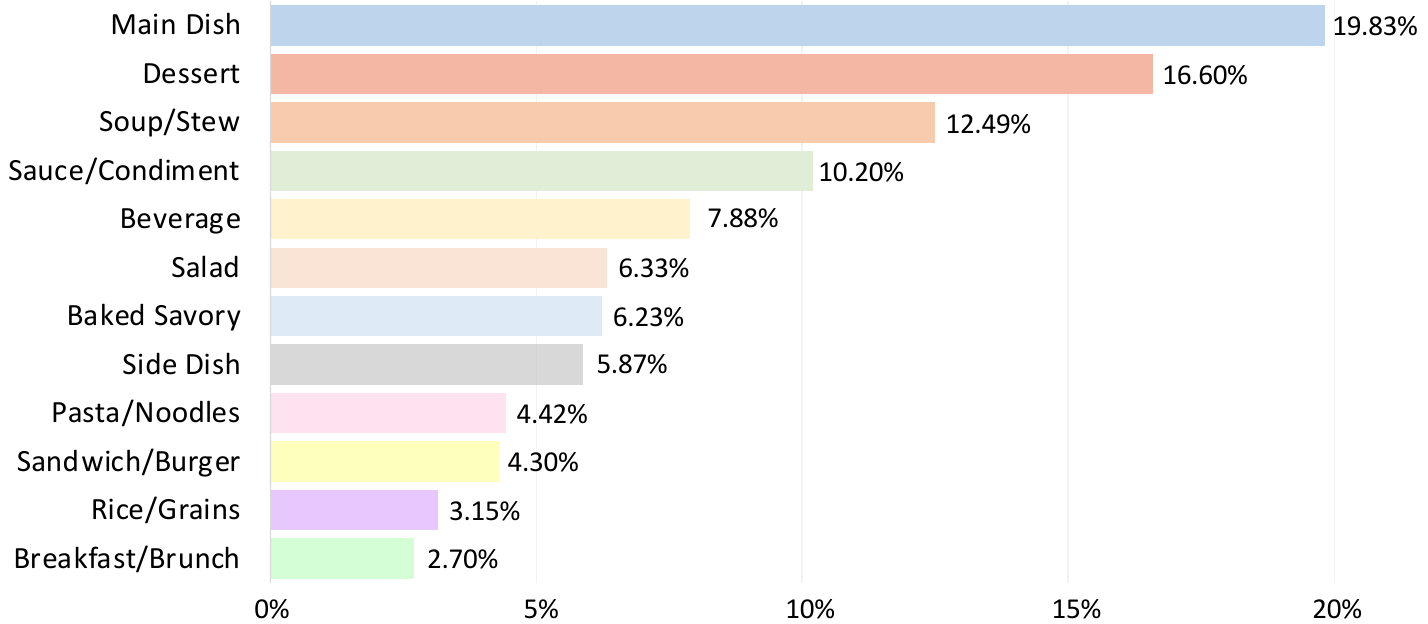}
    \caption{Category distribution of the ReGraph dataset.}
    \label{fig:category_distribution}
\vspace{-0.8em}
\end{figure}
\section{Dataset Construction}
\subsection{Dataset Source and Sampling}
The ReGraph dataset is constructed from Recipe1M \cite{recipe1m}. We sample 8,500 recipes from the training split and 1,500 recipes from the test split, resulting in a total of 10,000 recipes for structured graph annotation. As shown in Figure \ref{fig:category_distribution}, ReGraph follows the original category distribution of Recipe1M, ensuring a highly diverse and representative coverage of various cuisines. Each source recipe provides three primary textual inputs for graph construction: the recipe title, ingredient list, and step-by-step instructions.

\subsection{Automated Graph Generation}
Each ReGraph sample is produced through a two-stage generation pipeline. We represent each recipe as a structured graph composed of nodes and edges, where nodes correspond to entities and edges correspond to typed relations. Claude-Sonnet 4.5 \cite{claude45} first generates a procedural Recipe Reasoning Chain-of-Thought (RR-CoT) analysis, which sequentially decomposes recipe instructions to identify ingredient transformations and procedural ordering relationships, and extracts preliminary entities and relations from the recipe text. GPT-4o \cite{gpt4o} then performs schema normalization, converting these preliminary outputs into a consistent triple format with strict field-level alignment. Representative prompt excerpts and output examples are provided in the Appendix \ref{prompt}, while the complete prompt templates will be released together with the public dataset.

\textbf{Entity Extraction.}
The model extracts three primary entity types from each recipe: ingredient, action, and tool. Each entity is represented using five fields: \texttt{entity id}, \texttt{entity type}, \texttt{entity name}, \texttt{attributes} (e.g., `heated to boiling'), and \texttt{source step} for textual traceability. Whenever an action alters the state of an ingredient, the model must
generate a new ingredient entity with a unique \texttt{entity id} while
preserving the base entity name and recording the resulting state change
through the \texttt{attributes} field (e.g., an intermediate state such as
\texttt{mixture} with the attribute ``heated to boiling'' or ``boiled'').
Furthermore, the creation of any intermediate compound is explicitly
modeled as a new ingredient entity.

\textbf{Relation Extraction.}
The extracted entities are linked by three typed edges: \texttt{targ}, \texttt{dest}, and \texttt{followed by}. \texttt{targ} (Target) links an ingredient or other directly manipulated entity (head) to the action (tail) applied to it, capturing the input object of the action. \texttt{dest} (Destination) links an action (head) to its destination or output entity (tail), which may correspond to a tool, a transformed ingredient state, or a newly created combined product. By explicitly encoding these ordering constraints, \texttt{followed by} makes the procedural dependencies among cooking actions directly represented in the graph rather than left implicit in free-form recipe instructions. Each relation is represented with five fields: \texttt{head entity id}, \texttt{relation type}, \texttt{tail entity id}, \texttt{relation attributes} (e.g., `on top'), and \texttt{source step}.
A complete annotation example is provided in Appendix~\ref{case_study}. The example illustrates how ReGraph represents an entire cooking workflow through explicit entities and typed relations, representing ingredient state evolution through entity attributes, together with intermediate entities, tool interactions, and long-range procedural ordering. It complements the simplified example in Figure~\ref{kg_case} by demonstrating the full annotation schema used throughout the dataset.
\begin{table}[t]
\centering
\caption{Statistics of human revisions on the 1,500-recipe test split. Most revisions improve annotation consistency rather than correcting semantic errors. Revision categories are non-mutually exclusive; a recipe may
receive multiple types of corrections.}
\label{tab:revision_statistics}
\begin{tabular}{lc}
\toprule
\textbf{Revision Type} & \textbf{Percentage (\%)} \\
\midrule
Entity rename / normalize & 24.0 \\
Entity source-step realignment & 18.7 \\
Relation update reconnecting unlinked entities & 17.3 \\
Entity rename + attributes adjustment & 16.7 \\
Entity attributes refinement & 13.5 \\
Relation attributes update & 9.2 \\
Relation source-step realignment & 8.5 \\
Relation type correction & 4.9 \\
\bottomrule
\end{tabular}
\vspace{-0.8em}
\end{table}
\subsection{Quality Control and Human Validation}
Given the complexity of capturing precise ingredient-action transformations, we subject all 1,500 test-set recipes to mandatory multi-pass human validation by three annotators to ensure benchmark integrity. The validation targets semantic ambiguities and strict schema adherence. Annotators verify that the extracted graph structures faithfully reflect the original recipe text, checking entity integrity (e.g., raw vs.~processed chicken) and the separation of \texttt{entity name} from \texttt{attributes}. Revisions primarily standardize entity expressions, separate ingredient states from entity names, align source steps, and correct \texttt{targ}/\texttt{dest} assignments without altering the underlying cooking procedure. Validation also ensures that \texttt{followed by} explicitly represents ordering constraints and procedural dependencies among cooking actions. Table~\ref{tab:revision_statistics} summarizes the human revision statistics on the entire test split. Most revisions are normalization- and alignment-oriented rather than semantically substantive.
\begin{table}[h]
\centering
\setlength{\abovecaptionskip}{1pt}
\caption{
Annotation review consistency on 100 test recipes.
Annotators independently judged each entity and relation record in the same human-refined graph as correct or incorrect.
We report raw agreement and Cohen's $\kappa$ for each annotator pair; the final row reports three-annotator exact agreement and Fleiss' $\kappa$.
}
\label{tab:review_consistency}
\setlength{\tabcolsep}{4pt}
\begin{tabular}{lcc|cc}
\toprule
\multirow{2}{*}{\textbf{Annotators}}
& \multicolumn{2}{c|}{\textbf{Entity Records}}
& \multicolumn{2}{c}{\textbf{Relation Records}}
\\
\cmidrule(lr){2-3}
\cmidrule(l){4-5}
& \textbf{Agreement (\%)}
& \textbf{$\kappa$}
& \textbf{Agreement (\%)}
& \textbf{$\kappa$}
\\
\midrule
Annotator 1 \& 2
& 94.6 & 0.71
& 92.1 & 0.63
\\

Annotator 1 \& 3
& 95.1 & 0.73
& 92.8 & 0.66
\\

Annotator 2 \& 3
& 94.2 & 0.69
& 91.6 & 0.61
\\

\midrule
\textbf{All three annotators}
& \textbf{91.4} & \textbf{0.71}
& \textbf{88.7} & \textbf{0.63}
\\
\bottomrule
\end{tabular}
\end{table}

\begin{table}[h]
\centering
\setlength{\abovecaptionskip}{1pt}
\caption{Human audit of automatically constructed training annotations on 200 randomly sampled training recipes.}
\label{tab:train_audit}
\begin{tabular}{lc}
\toprule
\textbf{Metric} & \textbf{Score (\%)} \\
\midrule
Schema validity        & 100.0 \\
Entity correctness     & 93.0 \\
Relation correctness   & 89.5 \\
\bottomrule
\end{tabular}
\end{table}
\paragraph{Annotation Review Consistency.}
While the revision statistics describe the types of corrections made during human validation, we further examine the reliability of the resulting test annotations through an annotation review consistency study on 100 randomly sampled recipes. Each of the three annotators independently reviews the same graph produced after the preceding human-correction stage, using the source recipe and the ReGraph schema definition as references, and labels every entity and relation record as correct or incorrect. We report pairwise raw agreement and Cohen's $\kappa$, together with three-annotator exact agreement and Fleiss' $\kappa$. The high agreement in Table~\ref{tab:review_consistency} indicates that the corrected graph annotations are judged consistently across reviewers, providing additional evidence for the reliability of the test set. The recurring error patterns identified during test-set validation are subsequently incorporated into a semi-automatic cleaning pipeline for the training split, including entity normalization, source-step realignment, relation synchronization, and schema validation.

\paragraph{Training-Set Annotation Audit.}
Because the training annotations are cleaned semi-automatically rather than exhaustively reviewed, we additionally sample 200 training recipes for independent auditing by the same three annotators. Each annotator evaluates schema validity, entity correctness, and relation correctness, and the reported scores are averaged across annotators. As shown in Table~\ref{tab:train_audit}, the audited annotations achieve 100\% schema validity, 93\% entity correctness, and 89.5\% relation correctness. These results indicate that the cleaning pipeline produces consistently parsable and reliable training supervision, with the lower relation correctness reflecting the greater difficulty of relation assignment and action granularity.

\begin{table*}[htbp]
\centering
\small
\setlength{\abovecaptionskip}{1pt}
\caption{Total counts and average frequency per recipe of entities and relations in the ReGraph dataset.}
\label{tab:data_stats}
\resizebox{\textwidth}{!}{
\begin{tabular*}{\textwidth}{@{\extracolsep{\fill}}@{}lcccc@{\hspace{1.5em}}cccc@{}}
\toprule
\multirow{2}{*}{\textbf{Dataset Split}}
& \multicolumn{4}{c}{\textbf{Entities}}
& \multicolumn{4}{c}{\textbf{Relations}} \\
\cmidrule(lr){2-5}
\cmidrule(lr){6-9}
& Ingredient
& Tool
& Action
& \textbf{Total}
& \texttt{targ}
& \texttt{dest}
& \texttt{followed by}
& \textbf{Total} \\
\midrule
Training Set
& 142,175
& 28,102
& 100,705
& 270,982
& 149,649
& 96,266
& 83,695
& 329,610 \\

Test Set
& 25,065
& 4,981
& 17,745
& 47,791
& 27,343
& 18,227
& 15,871
& 61,441 \\

\midrule
\textbf{Total Count}
& \textbf{167,240}
& \textbf{33,083}
& \textbf{118,450}
& \textbf{318,773}
& \textbf{176,992}
& \textbf{114,493}
& \textbf{99,566}
& \textbf{391,051} \\

Avg. per Recipe
& 16.72
& 3.31
& 11.85
& 31.88
& 17.70
& 11.45
& 9.96
& 39.11 \\

\bottomrule
\end{tabular*}
}
\end{table*}
\subsection{Dataset Statistics}
The granularity of our annotation process generates a large volume of entities and relations per recipe. Table \ref{tab:data_stats} presents a breakdown of the dataset, detailing both the total accumulated counts across the training and test splits, and the average frequency of each entity (Ingredient, Tool, Action) and relation (\texttt{targ}, \texttt{dest}, \texttt{followed by}) type per recipe.

\begin{figure*}[t]
\centering
\includegraphics[width=1.0\textwidth]{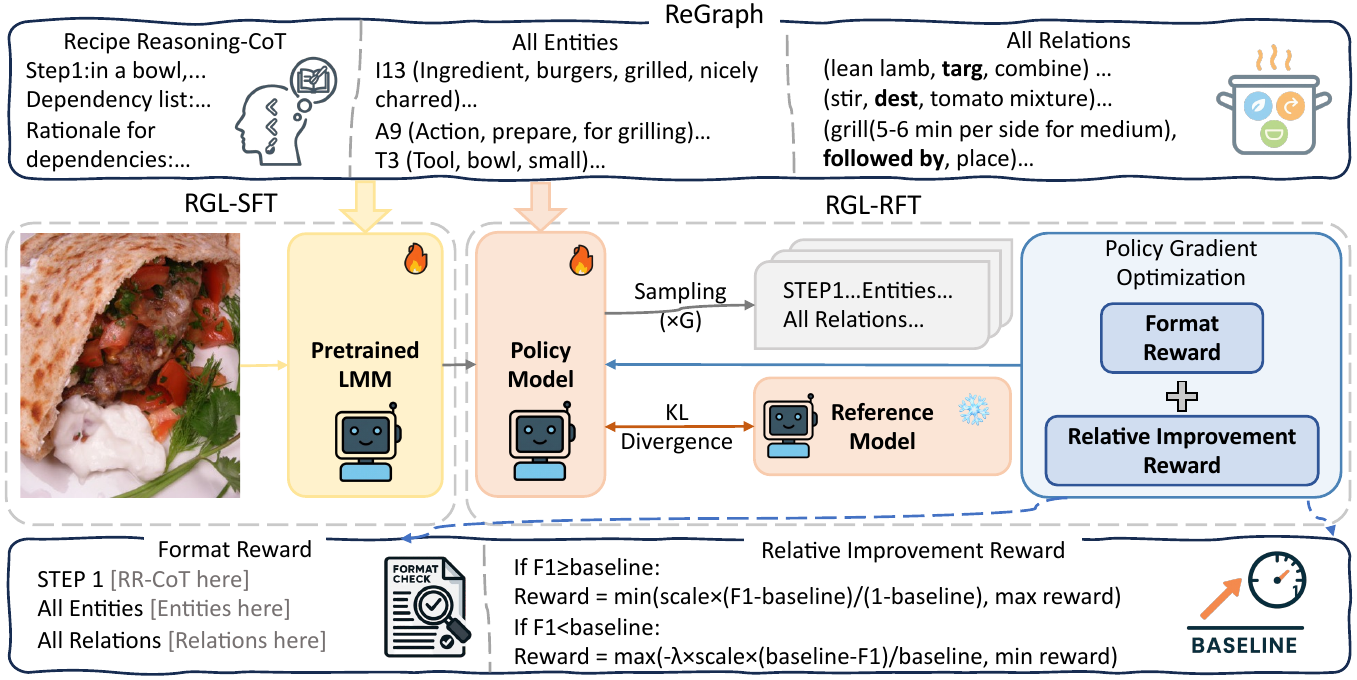} 
\caption{Overview of our two-stage Recipe Graph Learning (RGL) framework. RGL-SFT first teaches the model to generate structured graphs with Recipe Reasoning Chain-of-Thought (RR-CoT) data. RGL-RFT subsequently optimizes for graph-level semantic accuracy and structural validity using GRPO with proposed Relative Improvement Reward.}
\label{framework}
\end{figure*}

\section{Recipe Graph Learning Framework}
\label{method}
We propose Recipe Graph Learning (RGL), a two-stage training strategy for recipe graph generation from food images. As shown in Figure \ref{framework}, RGL-SFT guides the model to generate RR-CoT together with structured graph outputs, while RGL-RFT further improves reference-aligned graph quality and format compliance through reward-driven optimization.
\begin{table}[t]
\centering
\caption{Fixed training query used in RGL-SFT.}
\label{tab:training_query}
\footnotesize
\setlength{\tabcolsep}{6pt}
\renewcommand{\arraystretch}{1.05}

\begin{tabular}{p{0.95\linewidth}}
\toprule
\textbf{Training Query} \\
\midrule

\textbf{Input:} A food image.

\medskip

\textbf{Instruction:}

Infer a plausible cooking workflow for the depicted dish and generate a structured recipe graph.

Specifically,
\begin{itemize}[leftmargin=1.5em, itemsep=1pt, topsep=2pt, parsep=0pt]
    \item infer an ordered cooking procedure with procedural ordering constraints;
    \item generate an RR-CoT describing the reasoning process;
    \item extract entities (\texttt{ingredient}, \texttt{tool}, and \texttt{action});
    \item predict typed relations (\texttt{targ}, \texttt{dest}, and \texttt{followed by});
    \item output the recipe graph following the predefined schema.
\end{itemize}

\\
\bottomrule
\end{tabular}
\end{table}
\subsection{RGL-SFT: Recipe Graph Learning with Supervised Fine-Tuning}

To enable Large Multimodal Models to generate procedural recipe graphs from static food images, we first perform supervised fine-tuning on ReGraph. Since ingredient state changes, action--ingredient interactions, and procedural ordering are not directly observable from the final food appearance, supervising only the final graph provides limited explicit guidance for learning the intermediate procedural organization required to generate such structures. Therefore, the model is trained to generate an auxiliary reasoning trace together with the corresponding graph structure.

Each training instance in ReGraph contains an explicit Recipe Reasoning Chain-of-Thought (RR-CoT) trace. Rather than directly producing entities and relations, the model is supervised to first organize a plausible cooking procedure into a sequence of steps, identify the procedural ordering and dependencies among them, and then generate the corresponding structured graph. This intermediate trace provides procedural decomposition supervision that encourages the model to organize inferred cooking steps, ingredient state changes, intermediate products, and action-ingredient interactions before producing the final graph representation.

For each training instance, let $x$ denote the food image and let $q$ denote the fixed textual query shown in Table~\ref{tab:training_query}. The query instructs the model to infer a plausible cooking workflow from the input image and generate the corresponding structured recipe graph. Given the multimodal input pair $(x,q)$, the target sequence is represented as
$
o=[o_{\mathrm{reason}},o_{\mathrm{graph}}],
$
where $o_{\mathrm{reason}}$ denotes the RR-CoT trace and $o_{\mathrm{graph}}$ denotes the final recipe graph composed of structured entities and typed relations. The model is therefore optimized to generate an auxiliary procedural trace followed by a graph that is consistent with that trace and aligned with the ReGraph supervision. This formulation jointly models stepwise procedural organization and structured graph generation within a unified autoregressive framework.

The optimization objective of RGL-SFT is the standard negative log-likelihood loss over the training set $\mathcal{D}$:

\begin{equation}
\mathcal{L}_{\mathrm{SFT}}(\theta)
=
-
\mathbb{E}_{(x,q,o)\sim\mathcal{D}}
\left[
\frac{1}{|o|}
\sum_{t=1}^{|o|}
\log
\pi_\theta
(o_t \mid x,q,o_{<t})
\right].
\end{equation}

This objective maximizes the likelihood of generating both the RR-CoT trace and the corresponding graph conditioned on the multimodal input. As a result, RGL-SFT teaches the model the predefined graph schema while providing intermediate supervision for generating ingredient state representations, action sequences, intermediate products, and procedural ordering from visual inputs. During evaluation, only the generated graph is scored, while the RR-CoT serves solely as auxiliary supervision during supervised fine-tuning.
\subsection{RGL-RFT: Recipe Graph Learning with Reinforcement Fine-Tuning}

Although RGL-SFT equips the model with procedural organization and structured graph generation capabilities, token-level likelihood optimization does not directly encourage graph-level correctness. Small generation errors may propagate through the graph, leading to missing entities, incorrect relations, or inconsistent procedural ordering, even when most output tokens are predicted correctly. To further improve the quality of the generated recipe graphs, we perform a second-stage reinforcement fine-tuning based on Group Relative Policy Optimization (GRPO) \cite{GRPO}.

Unlike supervised learning, reinforcement learning directly optimizes task-level objectives through reward signals. In our setting, the model is encouraged to generate graphs with more accurate entities, more complete procedural relations, and schema-compliant outputs according to the reward functions defined below. Different from RGL-SFT, RGL-RFT applies semantic reinforcement optimization only to the generated recipe graph. Specifically, the Relative Improvement Reward (RIR) evaluates the entity and relation content of the graph without assessing the semantic quality of the RR-CoT, while the Format Reward (FR) only verifies that the reasoning section is present and structurally well formed. We adopt GRPO because it avoids training an additional value network and instead estimates advantages through relative comparisons among multiple candidate outputs generated for the same input, resulting in a memory-efficient optimization procedure.

For each training instance, let $x$ denote the food image and $q$ denote the fixed textual query. Given the multimodal input pair $(x,q)$, the previous policy $\pi_{\theta_{\mathrm{old}}}$ samples a group of $G$ candidate outputs,
$
\{o_1,o_2,\ldots,o_G\}.
$
Each candidate output is evaluated by the composite reward function defined below, producing rewards
$
\{R_1,R_2,\ldots,R_G\}.
$
Instead of optimizing these absolute rewards directly, GRPO computes a relative advantage by normalizing rewards within the sampled group:

\begin{equation}
\hat{A}_i
=
\frac
{R_i-\mathrm{mean}(\{R_1,\ldots,R_G\})}
{\mathrm{std}(\{R_1,\ldots,R_G\})}.
\end{equation}

This group-wise normalization makes optimization depend on the relative reward of candidate outputs generated from the same input rather than on their absolute reward values, thereby reducing the influence of reward-scale differences and providing a more informative learning signal.

The policy is then updated by comparing the likelihood of each sampled output under the current policy $\pi_\theta$ and the previous policy $\pi_{\theta_{\mathrm{old}}}$. Following PPO-style optimization, the policy ratio is clipped to the interval
$
[1-\epsilon,\,1+\epsilon]
$
to avoid excessively large policy updates. In addition, a KL-divergence regularization term weighted by $\beta$ constrains the updated policy from deviating excessively from the reference model $\pi_{\mathrm{ref}}$, preserving the generation capability acquired during supervised fine-tuning. The resulting optimization objective is

\begin{equation}
\begin{aligned}
\mathcal{J}_{\mathrm{GRPO}}(\theta)
&=
\mathbb{E}_{(x,q)\sim\mathcal D,\,
\{o_i\}\sim\pi_{\theta_{\mathrm{old}}}}
\Bigg[
\frac1G
\sum_{i=1}^{G}
\min
\Big(
\rho_i\hat A_i,
\\
&
\qquad
\mathrm{clip}
(\rho_i,1-\epsilon,1+\epsilon)
\hat A_i
\Big)
-
\beta
D_{\mathrm{KL}}
(
\pi_\theta
\|
\pi_{\mathrm{ref}}
)
\Bigg],
\end{aligned}
\end{equation}

\[
\rho_i
=
\frac
{\pi_\theta(o_i\mid x,q)}
{\pi_{\theta_{\mathrm{old}}}(o_i\mid x,q)},
\]

where $\epsilon$ denotes the clipping threshold and $\beta$ controls the strength of KL regularization.

\subsubsection{Relative Improvement Reward (RIR)}
\label{sec:rewards}

The primary objective of RGL-RFT is to directly optimize the reference-aligned structural quality of the generated recipe graph rather than token-level generation quality. The reward should therefore reflect how closely the predicted graph aligns with the procedural structure provided by the corresponding ReGraph annotation. To this end, we define graph correctness using exact entity and relation matching during reinforcement optimization.

A predicted entity is considered a True Positive (TP) only if its \texttt{entity type} and \texttt{entity name} field exactly match those of the corresponding reference entity. Likewise, a predicted relation is regarded as correct only when its \texttt{relation type} matches exactly and the entity types and names of both its head and tail entities satisfy the same matching criterion. This definition focuses on reference-aligned entity identification and typed procedural relations rather than superficial textual similarity. Ingredient-state attributes are not explicitly scored by the RIR reward. Nevertheless, improving ingredient entity identification can indirectly increase the number of correct joint name--state matches under the state-aware evaluation protocol.

During RGL-RFT, entity and relation F1 scores are computed independently for each sampled candidate against its corresponding training annotation. The F1 score is calculated based on the numbers of true positives (TP), false positives (FP), and false negatives (FN):

\[
F_1=\frac{2\cdot TP}{2\cdot TP+FP+FN}.
\]

These candidate-level scores are then used to compute the RIR reward.

Although F1 directly measures graph-matching quality, using raw F1 as the reinforcement reward treats all candidates on a uniform absolute scale, without accounting for the task-level performance already achieved by the supervised initialization or for the differing difficulty of entity and relation prediction. A baseline-relative transformation is therefore introduced to distinguish candidate scores above the SFT performance anchor from those below it and to provide a signed, task-specific reward signal.

To this end, we propose the \textbf{Relative Improvement Reward (RIR)}, which measures graph quality relative to the task-level performance of the supervised initialization rather than using the absolute F1 score directly. The key idea is to assign positive rewards to candidates that exceed the corresponding SFT baseline and softened negative rewards to candidates that fall below it, thereby producing a baseline-aware optimization signal.

Since entity prediction and relation prediction exhibit substantially different difficulty levels, we compute their rewards independently while sharing the same formulation. Specifically, two task-specific baselines, denoted by $b_{\mathrm{entity}}$ and $b_{\mathrm{relation}}$, are obtained from the macro-averaged per-sample F1 scores of the converged SFT model on the training set. Using a global task-specific baseline rather than per-sample reference scores provides a shared and consistent optimization anchor across training samples. It also allows entity and relation rewards to be calibrated relative to their respective task difficulty. Throughout reinforcement learning, RIR is computed using strict exact matching to provide deterministic reward signals, whereas the final evaluation additionally adopts exact-then-semantic matching.

Given an F1 score $F_1\in[0,1]$ and the corresponding baseline $b$, the Relative Improvement Reward is defined as

\begin{equation}
\mathrm{RIR}(F_1)=
\begin{cases}
\dfrac{F_1-b}{1-b}\cdot s,
& F_1\ge b,\\[2mm]
-\dfrac{b-F_1}{b}\cdot s\cdot\lambda,
& F_1<b,
\end{cases}
\end{equation}

where $s$ denotes a scaling factor and $\lambda=0.5$ is a penalty reduction factor. The positive branch measures the normalized margin above the task-specific SFT baseline, whereas the negative branch applies a softened penalty to predictions that fall below the baseline. Consequently, RIR distinguishes candidates above and below the task-level SFT performance anchor while reducing the penalty magnitude for lower-scoring candidates, which avoids excessively abrupt reward variation and calibrates entity and relation difficulty independently during policy optimization.

Finally, the reward is clipped to the fixed interval $[-0.5,1.0]$ to bound its magnitude, limit the influence of extreme reward values, and further improve optimization stability.
\subsubsection{Format Reward (FR)}

Besides graph-level correctness, valid recipe graph generation requires the output to conform to the predefined structural schema so that entities and relations can be reliably parsed for evaluation and reinforcement learning. To encourage schema-compliant generation, we introduce a lightweight \textbf{Format Reward (FR)} that evaluates whether the generated output satisfies the required output specification.

\begin{table}[h]
\centering
\caption{Structural constraints verified by the Format Reward (FR).}
\label{tab:format_reward}
\small
\begin{tabular}{lp{8cm}}
\toprule
\textbf{Constraint} & \textbf{Description} \\
\midrule
Reasoning section & The output contains an RR-CoT reasoning trace; its semantic content is not evaluated by FR. \\
Entity section & All graph entities are explicitly listed. \\
Relation section & All graph relations are explicitly listed. \\
Schema order & Sections follow the predefined output organization. \\
Record completeness & Entity and relation records contain all required fields for parsing. \\
\bottomrule
\end{tabular}
\end{table}
Rather than assessing semantic correctness, FR verifies only schema-level validity. Specifically, it checks whether the generated output contains all mandatory structural components, preserves the predefined organizational order, and provides complete entity and relation records that can be successfully parsed. FR also checks whether an RR-CoT section is structurally present, but does not evaluate the semantic correctness of its content. The structural constraints used by FR are summarized in Table~\ref{tab:format_reward}. A generated output receives a reward of 1 if all constraints are satisfied and 0 otherwise. Since these checks are deterministic and computationally inexpensive, FR provides an auxiliary supervision signal that discourages malformed outputs while complementing the graph-matching optimization performed by RIR.

Finally, the total reward for each candidate output $o$ is computed as a weighted combination of the entity-level and relation-level graph-matching rewards and the format-compliance reward:

\begin{equation}
R_{\mathrm{total}}(o)
=
\alpha \cdot
\mathrm{RIR}\!\left(F_{1}^{\mathrm{entity}}(o)\right)
+
\beta' \cdot
\mathrm{RIR}\!\left(F_{1}^{\mathrm{relation}}(o)\right)
+
\gamma \cdot \mathrm{FR}(o),
\end{equation}

where $F_{1}^{\mathrm{entity}}(o)$ and $F_{1}^{\mathrm{relation}}(o)$ denote the entity- and relation-level F1 scores of candidate output $o$, respectively. The reward weights are empirically set to $\alpha=1$, $\beta'=1$, and $\gamma=0.1$.

\section{Experiments}
\subsection{Experiment Settings}
\label{setting}

\noindent\textbf{Dataset and Baselines.}
We conduct experiments on the \textbf{ReGraph} dataset, using 8,500 samples for training both RGL-SFT and RGL-RFT and reserving 1,500 samples for evaluation. We adopt \textbf{Qwen3-VL-8B} \cite{qwen3} and \textbf{InternVL3-8B} \cite{internvl} as the base backbones to evaluate the generalizability of the proposed Recipe Graph Learning framework. The resulting supervised and reinforcement fine-tuned models are denoted as \textbf{Qwen3-VL-SFT}/\textbf{RFT} and \textbf{InternVL3-SFT}/\textbf{RFT}, respectively. For RGL-RFT, the RIR reference baselines $b_{\mathrm{entity}}$ and $b_{\mathrm{relation}}$ are initialized using the Entity F1 and Relation F1 scores achieved by the corresponding RGL-SFT model. This design allows the reward function to explicitly measure relative improvements over the supervised baseline for each task component.

\noindent\textbf{In-Context Learning Baselines.}
To assess the native capability of general-purpose LMMs for procedural recipe graph generation, we evaluate a diverse set of models without parameter updates, including the open-source models \textbf{InternVL3-8B} \cite{internvl}, \textbf{Qwen3-VL-8B}, and the larger \textbf{Qwen3-VL-32B} \cite{qwen3}, as well as the proprietary models \textbf{GPT-5.2} \cite{gpt52} and \textbf{Gemini 3.1 Pro} \cite{gemini31pro}. The open-source models are evaluated under four in-context learning settings: \textbf{zero-shot}, \textbf{1-shot}, \textbf{2-shot}, and \textbf{3-shot}. In the zero-shot setting, each model is prompted only with the task instructions and the target food image. In the few-shot settings, we provide $k \in \{1,2,3\}$ complete demonstrations, each consisting of a food image, its corresponding RR-CoT trace, and the target recipe graph.

\noindent\textbf{Implementation Settings.}
We perform parameter-efficient fine-tuning using LoRA \cite{lora} with a rank of $128$ within the ms-swift framework \cite{swift}. All experiments are conducted on a computing cluster equipped with four NVIDIA A100 GPUs, and AdamW is used as the optimizer throughout all training stages. To accommodate the lengthy Recipe Reasoning-CoT (RR-CoT) traces and structured recipe graph outputs, the maximum sequence length is set to $8{,}192$ tokens.

During supervised fine-tuning (\textbf{RGL-SFT}), both Qwen3-VL-8B and InternVL3-8B are trained for one epoch with a learning rate of $1 \times 10^{-4}$, a global batch size of $8$, and a warmup ratio of $0.05$. During reinforcement fine-tuning (\textbf{RGL-RFT}), the learning rate is reduced to $5 \times 10^{-6}$ while the global batch size remains $8$ to improve policy optimization stability. For GRPO exploration, we sample a group of $G=8$ candidate outputs for each query using nucleus sampling with top-$p=0.85$ and a temperature of $1.0$. The overall RGL-RFT reward is defined as a weighted combination of the entity reward, relation reward, and format reward, controlled by coefficients $\alpha$, $\beta'$, and $\gamma$, respectively. We set $\alpha=1$, $\beta'=1$, and $\gamma=0.1$. The format reward is assigned a relatively small weight because the preceding RGL-SFT stage already establishes strong structural compliance. It therefore serves primarily as a lightweight regularizer that prevents format collapse during reinforcement learning, while allowing the policy to focus on improving semantic entities and relations. A substantially larger format-reward weight could over-constrain the policy and reduce its capacity for semantic exploration.
\begin{table*}[t]
\centering
\footnotesize
\setlength{\abovecaptionskip}{1pt}
\caption{Comprehensive performance comparison on the ReGraph test set under the deterministic canonical matching protocol. We report Precision (P), Recall (R), and F1-score (\%) for entity and relation prediction. Ingredient entities are matched using their canonical names and retained action/state attributes, whereas action and tool entities are matched using canonical names only. Open-source LMMs are evaluated under zero-shot, 1-shot, 2-shot, and 3-shot settings, whereas proprietary LMMs are evaluated under zero-shot and 3-shot settings. ``\#Params'' denotes the parameter scale of the model backbone. The best result is shown in bold.}
\label{tab:main_results}
\begin{tabular}{l|c|c|ccc|ccc}
\toprule
\multirow{2}{*}{\textbf{Method}}
& \multirow{2}{*}{\textbf{\#Params}}
& \multirow{2}{*}{\textbf{Setting}}
& \multicolumn{3}{c|}{\textbf{Entity}}
& \multicolumn{3}{c}{\textbf{Relation}}
\\
\cmidrule(lr){4-6}
\cmidrule(l){7-9}
& & &
\textbf{P} & \textbf{R} & \textbf{F1}
& \textbf{P} & \textbf{R} & \textbf{F1}
\\
\midrule

\multicolumn{9}{c}{\textit{Existing General-Purpose LMMs}}
\\
\midrule

\multirow{4}{*}{InternVL3 \cite{internvl}}
& \multirow{4}{*}{8B}
& 0-shot
& 10.79 & 5.03 & 6.92
& 0.60 & 0.34 & 0.44
\\

& & 1-shot
& 18.79 & 12.76 & 15.25
& 2.36 & 1.37 & 1.72
\\

& & 2-shot
& 19.96 & 13.76 & 16.33
& 2.72 & 1.84 & 2.19
\\

& & 3-shot
& 20.48 & 14.01 & 16.66
& 3.29 & 2.24 & 2.66
\\
\midrule

\multirow{4}{*}{Qwen3-VL \cite{qwen3}}
& \multirow{4}{*}{8B}
& 0-shot
& 19.65 & 5.86 & 9.15
& 1.90 & 0.88 & 1.20
\\

& & 1-shot
& 26.20 & 15.44 & 19.41
& 4.29 & 2.57 & 3.20
\\

& & 2-shot
& 24.52 & 13.49 & 17.39
& 3.66 & 2.03 & 2.59
\\

& & 3-shot
& 23.56 & 14.13 & 17.76
& 3.90 & 2.36 & 2.93
\\
\midrule

\multirow{4}{*}{Qwen3-VL \cite{qwen3}}
& \multirow{4}{*}{32B}
& 0-shot
& 24.24 & 8.85 & 13.12
& 2.27 & 1.31 & 1.66
\\

& & 1-shot
& 21.25 & 17.39 & 19.13
& 3.08 & 2.57 & 2.80
\\

& & 2-shot
& 21.83 & 18.21 & 19.85
& 3.60 & 3.04 & 3.31
\\

& & 3-shot
& 19.88 & 15.86 & 17.65
& 3.77 & 3.05 & 3.37
\\
\midrule

\multirow{2}{*}{GPT-5.2 \cite{gpt52}}
& \multirow{2}{*}{N/A}
& 0-shot
& 21.18 & 12.25 & 15.62
& 3.07 & 2.15 & 2.52
\\

& & 3-shot
& 19.48 & 15.25 & 17.13
& 3.83 & 2.96 & 3.34
\\
\midrule

\multirow{2}{*}{Gemini 3.1 Pro \cite{gemini31pro}}
& \multirow{2}{*}{N/A}
& 0-shot
& 22.48 & 13.55 & 16.99
& 3.36 & 2.35 & 2.77
\\

& & 3-shot
& 20.87 & 16.10 & 18.20
& 4.16 & 3.21 & 3.62
\\
\midrule

\multicolumn{9}{c}{\textit{Recipe Graph Learning Framework}}
\\
\midrule

\rowcolor{gray!10}
InternVL3-SFT
&
& SFT
& 29.44 & 26.27 & 27.75
& 7.07 & 6.49 & 6.78
\\

\rowcolor{gray!10}
InternVL3-RFT
& \multirow{-2}{*}{8B}
& RFT
& 30.66 & \textbf{29.07} & 29.28
& 8.07 & 7.58 & 7.82
\\
\midrule

\rowcolor{gray!10}
Qwen3-VL-SFT
&
& SFT
& 30.66 & 25.72 & 26.75
& 8.12 & 6.36 & 7.13
\\

\rowcolor{gray!10}
Qwen3-VL-RFT
& \multirow{-2}{*}{8B}
& RFT
& \textbf{33.54} & 28.70 & \textbf{30.93}
& \textbf{9.29} & \textbf{8.03} & \textbf{8.62}
\\
\bottomrule
\end{tabular}
\end{table*}

\subsection{Evaluation Protocol}
\label{eval_protocol}

To evaluate the generated recipe graphs, we first parse the entity and relation sections from each model output while excluding the Recipe Reasoning Chain-of-Thought (RR-CoT) trace from metric computation. We report standard \textbf{Precision (P)}, \textbf{Recall (R)}, and \textbf{F1-score (F1)} for entity and relation prediction. Metrics are computed independently for each test instance and then macro-averaged over the test set. The primary evaluation adopts a deterministic, schema-aware canonical matching protocol. All training-derived vocabularies and mappings are constructed exclusively from the 8,500 training samples and frozen before test-set evaluation.

\noindent\textbf{Canonical Vocabulary Construction.}
Ingredient names are normalized using the ingredient vocabulary and synonym mapping adopted by Inverse Cooking \cite{inverse}. For ingredient attributes, we retain only explicit culinary operations and meaningful resulting states that describe ingredient transformations. Attribute strings are lowercased and whitespace-normalized, while result-state grammar is preserved rather than reduced to the corresponding action verb. For example, ``finely chopped'' is normalized to \texttt{chopped}, while ongoing and completed states such as \texttt{boiling} and \texttt{boiled} remain distinct. When multiple states are present, their original order is retained, as in \texttt{drained|rinsed}. Quantities, units, time, temperature, location, purpose, composition, and other non-state modifiers are removed; for example, ``1 tablespoon, melted'' retains only \texttt{melted}.

Action and tool vocabularies are constructed from the ReGraph training annotations. Action expressions are mapped using exact labels, explicit aliases, and the longest valid action verb occurring in the expression. For example, ``bring to boil'' is mapped to \texttt{boil}. High-frequency training actions not covered by the initial action list are retained as additional canonical labels. Tool expressions are lowercased and normalized using separator normalization, conservative singularization, and explicit synonym mappings. For example, ``mixing bowl'' is mapped to \texttt{bowl}. After variant aggregation, canonical action and tool labels with an aggregate training frequency below 10 are removed.

Applying this procedure to the training split reduces 12,626 unique raw ingredient-attribute strings to 246 retained action/state labels, 3,080 unique raw action expressions to 277 canonical actions, and 4,171 unique raw tool expressions to 137 canonical tools. The complete normalization vocabularies and mappings will be released with the ReGraph evaluation toolkit.

\noindent\textbf{Entity True-Positive Criteria.}
Entity matching is type-specific. An ingredient entity is counted as a true positive only when its entity type, canonical name, and retained action/state attributes match the reference. Thus, entities with the same ingredient name but different states, such as \texttt{mixture(melted)} and \texttt{mixture(boiled)}, are treated as distinct, while quantities and other non-state attributes are ignored. Action and tool entities are matched using only their entity type and canonical name; their attributes are excluded from scoring.

\noindent\textbf{Relation True-Positive Criteria.}
A relation is counted as a true positive only when its relation type matches and both endpoints satisfy the corresponding entity-matching criteria. Action and tool endpoints are matched by canonical name, whereas ingredient endpoints must additionally match their retained action/state attributes. Entity IDs are ignored because they are sample-internal identifiers without semantic meaning.

\noindent\textbf{Supplementary Semantic Verification.}
Although canonical normalization resolves most lexical variation, a fixed vocabulary may not cover every semantically interchangeable expression, particularly among cooking actions such as ``mix'' and ``combine.'' We therefore conduct an additional semantic-relaxation analysis using Gemini 3.1 Pro \cite{gemini31pro} and Qwen3-32B \cite{qwen3} as two independent semantic verifiers. Semantic verification is applied only to prediction--reference pairs that remain unmatched after canonical normalization and performs a restricted equivalence judgment rather than open-ended similarity assessment. The verifier may accept residual name-level equivalence only when the two expressions are functionally interchangeable in the culinary context without changing action specificity, tool semantics, ingredient identity or state, relation type, or procedural-ordering meaning. It cannot override ingredient-state distinctions. Results based on this LLM-assisted semantic relaxation are reported separately in the subsequent analysis experiments and are not included in the primary result table. The complete semantic-verification prompts will be released together with the ReGraph evaluation toolkit.
\subsection{Performance Comparison}
\label{sec:performance}
Table~\ref{tab:main_results} reports recipe graph generation results under our primary deterministic canonical matching protocol; results under supplementary LLM-assisted semantic matching are reported in Table~\ref{tab:canonical_semantic_full}. The reported Entity and Relation scores measure alignment between the generated graph and the annotated reference workflow.

\noindent\textbf{In-Context Learning Results.}
Under the deterministic canonical matching protocol, general-purpose LMMs show limited ability to produce reference-aligned recipe graphs. Additional demonstrations improve schema conformity but yield inconsistent gains across scales and shot settings: InternVL3-8B improves progressively, whereas Qwen3-VL-8B peaks at 1-shot and Qwen3-VL-32B at 2-shot. Scaling from 8B to 32B yields no consistent improvement across entity and relation prediction. Relation generation is especially hard: the strongest baseline, Gemini 3.1 Pro (3-shot), reaches only 3.62\% Relation F1. These low scores do not mean the generated procedures are implausible; rather, free-form outputs contain little procedural information that can be consistently formalized into reference-aligned action targets, intermediate states, and ordering relations. In-context examples help models imitate the schema but cannot teach the detailed image-to-graph mapping, motivating task-specific supervision and graph-level reinforcement fine-tuning.

\noindent\textbf{Effectiveness of RGL.}
Both RGL-SFT and RGL-RFT substantially outperform all in-context baselines, and the improvement is consistent across two backbones. On Qwen3-VL, RGL-SFT achieves 26.75\% Entity F1 and 7.13\% Relation F1, outperforming the best proprietary baseline in relation prediction. This shows that supervised training equips the model to recover culinary entities and infer procedural dependencies from static images.
\begin{table}[t]
\centering
\caption{
Fine-grained breakdown of entity and relation generation under the deterministic canonical matching protocol.
We report F1-score (\%) for each entity and relation type.
Ingredient entities are matched using canonical names and retained action/state attributes, whereas action and tool entities are matched using canonical names only.
$\Delta$ denotes the improvement of RFT over SFT.
}
\label{tab:breakdown_all}

\small
\setlength{\tabcolsep}{5pt}
\begin{tabular}{lccc | lccc}
\toprule
\textbf{Entity Type}
& \textbf{SFT}
& \textbf{RFT}
& \textbf{$\Delta$}
&
\textbf{Relation Type}
& \textbf{SFT}
& \textbf{RFT}
& \textbf{$\Delta$}
\\
\midrule

Ingredient + State
& 17.51
& \textbf{21.40}
& +3.89
&
\texttt{targ}
& 6.76
& \textbf{8.17}
& +1.41
\\

Tool
& 24.42
& \textbf{28.18}
& +3.76
&
\texttt{dest}
& 6.56
& \textbf{7.93}
& +1.37
\\

Action
& 31.64
& \textbf{36.06}
& +4.42
&
\texttt{followed by}
& 8.50
& \textbf{10.28}
& +1.78
\\

\bottomrule
\end{tabular}
\end{table}

Table~\ref{tab:breakdown_all} breaks down the improvement by entity and relation type under the canonical matching protocol. RFT improves every category over SFT. Among relations, \texttt{followed by} gains the most (+1.78 F1), indicating that reinforcement fine-tuning is especially effective at recovering reference-aligned procedural ordering between cooking actions, while the consistent gains on \texttt{targ} (+1.41) and \texttt{dest} (+1.37) show strengthened action-target and destination modeling. Among entities, the state-aware Ingredient score improves by +3.89 yet remains the lowest in absolute terms (21.40 F1), confirming that fine-grained ingredient-state capture is the hardest dimension even after reinforcement fine-tuning.
\begin{table*}[h]
\centering
\footnotesize
\caption{
Comparison between deterministic canonical matching and supplementary LLM-assisted semantic matching on the ReGraph test set.
We report Precision (P), Recall (R), and F1-score (\%) for entity and relation prediction.
Canonical Match follows the primary schema-aware evaluation protocol.
Semantic matching further applies localized equivalence verification to prediction--reference pairs that remain unmatched after canonical normalization, using Gemini 3.1 Pro and Qwen3-32B independently.
}
\label{tab:canonical_semantic_full}

\resizebox{\textwidth}{!}{
\begin{tabular}{l|c|c|ccc|ccc|ccc|ccc|ccc|ccc}
\toprule

\multirow{3}{*}{\textbf{Method}}
& \multirow{3}{*}{\textbf{\#Params}}
& \multirow{3}{*}{\textbf{Setting}}
& \multicolumn{6}{c|}{\textbf{Canonical Match}}
& \multicolumn{6}{c|}{\textbf{Semantic Match (Gemini 3.1 Pro)}}
& \multicolumn{6}{c}{\textbf{Semantic Match (Qwen3-32B)}}
\\

\cmidrule(lr){4-9}
\cmidrule(lr){10-15}
\cmidrule(l){16-21}

& &
& \multicolumn{3}{c|}{\textbf{Entity}}
& \multicolumn{3}{c|}{\textbf{Relation}}
& \multicolumn{3}{c|}{\textbf{Entity}}
& \multicolumn{3}{c|}{\textbf{Relation}}
& \multicolumn{3}{c|}{\textbf{Entity}}
& \multicolumn{3}{c}{\textbf{Relation}}
\\

\cmidrule(lr){4-6}
\cmidrule(lr){7-9}
\cmidrule(lr){10-12}
\cmidrule(lr){13-15}
\cmidrule(lr){16-18}
\cmidrule(l){19-21}

& &
& \textbf{P} & \textbf{R} & \textbf{F1}
& \textbf{P} & \textbf{R} & \textbf{F1}
& \textbf{P} & \textbf{R} & \textbf{F1}
& \textbf{P} & \textbf{R} & \textbf{F1}
& \textbf{P} & \textbf{R} & \textbf{F1}
& \textbf{P} & \textbf{R} & \textbf{F1}
\\

\midrule

\multicolumn{21}{c}{\textit{Existing General-Purpose LMMs}}
\\
\midrule

\multirow{2}{*}{InternVL3 \cite{internvl}}
& \multirow{2}{*}{8B}
& 0-shot
& 10.79 & 5.03 & 6.92
& 0.60 & 0.34 & 0.44
& 12.71 & 6.28 & 8.41
& 1.15 & 0.63 & 0.82
& 12.44 & 6.27 & 8.34
& 1.21 & 0.68 & 0.87
\\

& & 3-shot
& 20.48 & 14.01 & 16.66
& 3.29 & 2.24 & 2.66
& 24.12 & 17.50 & 20.25
& 6.32 & 4.15 & 5.01
& 24.11 & 17.54 & 20.31
& 6.22 & 4.08 & 4.93
\\

\midrule

\multirow{2}{*}{Qwen3-VL \cite{qwen3}}
& \multirow{2}{*}{8B}
& 0-shot
& 19.65 & 5.86 & 9.15
& 1.90 & 0.88 & 1.20
& 23.15 & 7.32 & 11.12
& 3.65 & 1.63 & 2.25
& 23.07 & 7.23 & 11.01
& 3.74 & 1.68 & 2.32
\\

& & 3-shot
& 23.56 & 14.13 & 17.76
& 3.90 & 2.36 & 2.93
& 27.75 & 17.65 & 21.58
& 7.48 & 4.38 & 5.52
& 27.91 & 17.68 & 21.65
& 7.42 & 4.31 & 5.45
\\

\midrule

\multicolumn{21}{c}{\textit{Recipe Graph Learning Framework}}
\\
\midrule

\rowcolor{gray!10}
Qwen3-VL-SFT
& 8B
& SFT
& 30.66 & 25.72 & 26.75
& 8.12 & 6.36 & 7.13
& 36.12 & 29.62 & 32.51
& 15.58 & 11.78 & 13.41
& 35.91 & 29.57 & 32.43
& 15.62 & 11.84 & 13.47
\\

\rowcolor{gray!10}
Qwen3-VL-RFT
& 8B
& RFT
& \textbf{33.54} & \textbf{28.70} & \textbf{30.93}
& \textbf{9.29} & \textbf{8.03} & \textbf{8.62}
& \textbf{39.51} & \textbf{35.84} & \textbf{37.59}
& \textbf{17.82} & \textbf{14.88} & \textbf{16.22}
& \textbf{39.27} & \textbf{36.14} & \textbf{37.64}
& \textbf{17.64} & \textbf{14.83} & \textbf{16.11}
\\

\bottomrule
\end{tabular}
}
\end{table*}

\begin{table}[t]
\centering
\caption{
Effect of ingredient-state matching on entity and relation evaluation.
Name-only matching considers only canonical ingredient names, whereas state-aware matching additionally requires retained ingredient action/state attributes to match.
Action and tool matching remains unchanged under both settings.
}
\label{tab:state_matching_analysis}

\small
\setlength{\tabcolsep}{5pt}
\begin{tabular}{l|cc|cc}
\toprule
\multirow{2}{*}{\textbf{Metric}}
& \multicolumn{2}{c|}{\textbf{Name Only}}
& \multicolumn{2}{c}{\textbf{State Aware}}
\\
\cmidrule(lr){2-3}
\cmidrule(l){4-5}
& \textbf{SFT} & \textbf{RFT}
& \textbf{SFT} & \textbf{RFT}
\\
\midrule

Ingredient Entity F1
& 29.97
& \textbf{34.66}
& 17.51
& \textbf{21.40}
\\

\texttt{targ} F1
& 11.18
& \textbf{13.52}
& 6.76
& \textbf{8.17}
\\

\texttt{dest} F1
& 9.48
& \textbf{11.46}
& 6.56
& \textbf{7.93}
\\

\bottomrule
\end{tabular}
\end{table}
\noindent\textbf{Robustness across Evaluation Settings.}
We verify that the observed improvements do not depend on the matching protocol or the choice of semantic verifier. As shown in Table~\ref{tab:canonical_semantic_full}, scores increase consistently from Canonical Match to Semantic Match because the localized equivalence check recovers valid lexical variants (e.g., ``mix'' vs.\ ``combine'') that are not covered by the fixed canonical vocabulary; Canonical Match therefore provides a conservative evaluation. Crucially, the SFT--RFT ordering remains unchanged across all three settings. Qwen3-VL-RFT improves Entity F1 from 26.75 to 30.93 under Canonical Match, and similar gains are observed with both semantic verifiers, from 32.51 to 37.59 using Gemini 3.1 Pro and from 32.43 to 37.64 using Qwen3-32B, with an analogous pattern for relation prediction. Moreover, the two semantic verifiers produce nearly identical results, such as 37.59 and 37.64 Entity F1 for RFT, indicating that evaluator variation does not affect the overall conclusions. These results suggest that the gains arise from improved structured recipe graph generation rather than from relaxed matching or reliance on a particular LLM-based evaluator.

\noindent\textbf{Fine-Grained Ingredient-State Capture.}
A central goal of ReGraph is to make ingredient state transitions explicit and measurable. Table~\ref{tab:state_matching_analysis} isolates this dimension by comparing name-only matching, which requires only the canonical ingredient name, with state-aware matching, which additionally requires the retained action/state attribute. Across all metrics, state-aware scores are markedly lower than their name-only counterparts. For RFT, Ingredient Entity F1 decreases from 34.66 to 21.40, \texttt{targ} F1 from 13.52 to 8.17, and \texttt{dest} F1 from 11.46 to 7.93. This confirms that recovering fine-grained ingredient states is substantially more difficult than identifying the ingredients themselves, a distinction that conventional name-level evaluation cannot reveal. RFT improves state-aware Ingredient F1 from 17.51 to 21.40, indicating improved joint alignment between ingredient entities and their associated states. However, state-aware performance remains substantially lower than name-only matching, highlighting the difficulty of fine-grained ingredient-state modeling and motivating the explicit procedural representation provided by ReGraph.

\subsection{Explicit Procedural Structure in Generated Recipes}
\label{sec:recipe_gen_limits}
Recent image-to-recipe approaches achieve competitive scores on text-generation metrics such as SacreBLEU~\cite{sacrebleu} and ROUGE-L~\cite{lin2004rouge}, but these measure lexical similarity and do not reveal whether the generated instructions encode an explicit, consistently structured cooking process. To examine this, we convert the recipes generated by representative methods into graphs using the same extraction pipeline and evaluate them against the ReGraph annotations under the canonical matching protocol (Table~\ref{tab:recipe_generation_graph}). This does not assume a unique hidden process; it measures how consistently free-form instructions can be formalized into reference-aligned entities and relations.

Despite competitive text-generation performance (e.g., ROUGE-L above 42 for RecipeRAG and SGRG), these methods yield limited procedural structure: Entity F1 ranges from 10.27\% to 23.55\%, but Relation F1 remains between 0.67\% and 3.30\%. Notably, RecipeRAG attains the highest entity recall (32.03) yet only 3.30\% Relation F1, showing that models may mention plausible ingredients while failing to organize them into recoverable action targets, intermediate states, and ordering relations. RGL, which generates graphs directly, reaches 30.93\% Entity and 8.62\% Relation F1 under the same protocol. These results expose a substantial gap between fluent recipe text and reference-aligned procedural structure, motivating ReGraph as explicit supervision and RGL as a framework for generating fine-grained cooking structure from images.

\begin{table*}[t]
\centering
\caption{Explicit procedural structure extracted from recipes generated by representative image-to-recipe approaches. Generated recipes are converted into graphs using a common extraction pipeline and evaluated under the deterministic canonical matching protocol.}
\label{tab:recipe_generation_graph}

\begin{tabular}{l|cc|ccc|ccc}
\toprule

\multirow{2}{*}{\textbf{Method}}
&
\multicolumn{2}{c|}{\textbf{Recipe Generation}}
&
\multicolumn{3}{c|}{\textbf{Entity}}
&
\multicolumn{3}{c}{\textbf{Relation}}
\\

\cmidrule(lr){2-3}
\cmidrule(lr){4-6}
\cmidrule(l){7-9}

&
\textbf{SacreBLEU}
&
\textbf{ROUGE-L}
&
\textbf{P}
&
\textbf{R}
&
\textbf{F1}
&
\textbf{P}
&
\textbf{R}
&
\textbf{F1}
\\

\midrule

Inverse Cooking \cite{inverse}
& 3.64 & 32.41
& 11.50 & 10.08 & 10.27
& 0.67 & 0.67 & 0.67
\\

FoodLMM \cite{foodlmm}
& 3.96 & 27.94
& 11.95 & 11.37 & 11.71
& 0.71 & 0.75 & 0.73
\\

SDRA \cite{liu2025retrieval}
& 5.81 & 37.58
& 15.57 & 17.77 & 16.32
& 1.06 & 1.10 & 1.08
\\

RecipeRAG \cite{reciperag}
& 9.53 & 42.10
& 15.64 & \textbf{32.03} & 21.33
& 2.05 & 5.69 & 3.30
\\

SGRG \cite{liu2026enhancing}
& 8.27 & 42.65
& 18.98 & 31.93 & 23.55
& 2.52 & 2.85 & 2.68
\\

\midrule

\rowcolor{gray!10}
RGL-SFT (Ours)
& -- & --
& 30.66 & 25.72 & 26.75
& 8.12 & 6.36 & 7.13
\\

\rowcolor{gray!10}
\textbf{RGL-RFT (Ours)}
& -- & --
& \textbf{33.54}
& 28.70
& \textbf{30.93}
& \textbf{9.29}
& \textbf{8.03}
& \textbf{8.62}
\\

\bottomrule
\end{tabular}
\end{table*}

\subsection{Ablation Study}
\label{sec:ablation}

To verify the design choices in our framework, we conduct ablation studies on reward function design, component necessity, and hyperparameter configuration, with all experiments conducted on Qwen3-VL-8B.
\begin{table}[h]
\centering
\setlength{\abovecaptionskip}{1pt}
\caption{Ablation on reward design. We compare the proposed Relative Improvement Reward (RIR) against directly using raw F1 as the reward, and evaluate the effect of removing the Format Reward (FR). All results are reported under the deterministic canonical matching protocol.}
\label{tab:ablation_reward_design}
\setlength{\tabcolsep}{3.5pt}
\begin{tabular}{lcc|ccc|ccc}
\hline
\textbf{Setting}
& \textbf{RIR}
& \textbf{FR}
& \multicolumn{3}{c|}{\textbf{Entity}}
& \multicolumn{3}{c}{\textbf{Relation}}
\\
& & & P & R & F1 & P & R & F1
\\
\hline

Raw F1 reward ($R=F_1$)
& 
& $\checkmark$
& 32.29
& 28.49
& 30.13
& 7.55
& 7.04
& 7.23
\\

w/o FR
& $\checkmark$
&
& 31.61
& 27.30
& 29.30
& 6.59
& 5.82
& 6.18
\\

\textbf{Full reward design (Ours)}
& $\checkmark$
& $\checkmark$
& \textbf{33.54}
& \textbf{28.70}
& \textbf{30.93}
& \textbf{9.29}
& \textbf{8.03}
& \textbf{8.62}
\\

\hline
\end{tabular}
\end{table}

\noindent\textbf{Effectiveness of Reward Design.}
We evaluate our reward design by replacing the Relative Improvement Reward (RIR) with the raw F1 score and by removing the Format Reward (FR). As shown in Table~\ref{tab:ablation_reward_design}, RIR outperforms the raw F1 reward on both entity and relation prediction, achieving 30.93 versus 30.13 Entity F1 and 8.62 versus 7.23 Relation F1, with a larger margin on the more challenging relation prediction. By softening penalties for below-baseline candidates, bounding the reward range, and calibrating entity and relation difficulty independently, RIR provides a more effective optimization signal than raw F1. Removing FR causes a substantially larger performance drop, particularly for relations, where F1 decreases from 8.62 to 6.18, indicating that format-level supervision contributes to graph-level prediction performance, especially relation generation.
\begin{table}[h]
\centering
\setlength{\abovecaptionskip}{1pt}
\caption{Ablation on the effect of RR-CoT supervision during supervised fine-tuning. All results are reported under the deterministic canonical matching protocol.}
\label{tab:ablation_cot}
\setlength{\tabcolsep}{3.5pt}
\begin{tabular}{lcccccc}
\hline
\multicolumn{1}{l|}{\multirow{2}{*}{\textbf{Settings}}}
& \multicolumn{3}{c|}{\textbf{Entity}}
& \multicolumn{3}{c}{\textbf{Relation}} \\
\multicolumn{1}{l|}{}
& P & R & \multicolumn{1}{c|}{F1}
& P & R & F1 \\ 
\hline

\multicolumn{1}{l|}{w/o RR-CoT}
& 25.03
& \textbf{26.01}
& \multicolumn{1}{c|}{25.29}
& 6.26
& \textbf{6.37}
& 6.34 \\

\multicolumn{1}{l|}{w/ RR-CoT (Ours)}
& \textbf{30.66}
& 25.72
& \multicolumn{1}{c|}{\textbf{26.75}}
& \textbf{8.12}
& 6.36
& \textbf{7.13} \\

\hline
\end{tabular}
\end{table}
\noindent\textbf{Impact of Chain-of-Thought (CoT) Reasoning.}
Table~\ref{tab:ablation_cot} shows that RR-CoT supervision improves both Entity and Relation F1, with gains mainly driven by higher precision while recall remains largely unchanged. This suggests that the intermediate reasoning traces help the model generate more accurate entities and procedural relations with fewer false positives.
\begin{table}[h]
\centering
\caption{Ablation study on the scaling factor $s$ in RIR. We report Precision (P), Recall (R), and F1-score (\%) on the test set under the deterministic canonical matching protocol.}
\label{tab:ablation_scale}
\setlength{\tabcolsep}{3.5pt}
\begin{tabular}{c|ccc|ccc}
\hline
\multirow{2}{*}{\textbf{Scale ($s$)}}
& \multicolumn{3}{c|}{\textbf{Entity}}
& \multicolumn{3}{c}{\textbf{Relation}}
\\
& P & R & F1
& P & R & F1
\\
\hline

1
& 32.56
& 27.58
& 29.86
& 8.46
& 7.27
& 7.82
\\

\textbf{2}
& \textbf{33.54}
& \textbf{28.70}
& \textbf{30.93}
& \textbf{9.29}
& \textbf{8.03}
& \textbf{8.62}
\\

3
& 27.09
& 28.28
& 27.67
& 6.86
& 7.27
& 7.06
\\

\hline
\end{tabular}
\end{table}


\noindent\textbf{Impact of the Scaling Factor.}
We independently train RGL-RFT with $s \in \{1,2,3\}$, while fixing the reward weights to $\alpha=1$, $\beta'=1$, and $\gamma=0.1$. All other training and evaluation settings remain unchanged. As shown in Table~\ref{tab:ablation_scale}, $s=2$ achieves the best entity and relation performance among the evaluated settings.
\begin{table}[htbp]
\centering
\caption{Ablation study on the entity and relation reward weights ($\alpha$ for Entity and $\beta'$ for Relation) in RGL-RFT. The format reward weight $\gamma$ is fixed at 0.1 across all experiments. All results are reported under the deterministic canonical matching protocol. Best scores are shown in bold.}
\label{tab:reward_weights}
\setlength{\tabcolsep}{3.5pt}
\begin{tabular}{cc|ccc|ccc}
\toprule
\multicolumn{2}{c|}{\textbf{Reward Weights}}
& \multicolumn{3}{c|}{\textbf{Entity}}
& \multicolumn{3}{c}{\textbf{Relation}}
\\

\textbf{$\alpha$}
& \textbf{$\beta'$}
& \textbf{P}
& \textbf{R}
& \textbf{F1}
& \textbf{P}
& \textbf{R}
& \textbf{F1}
\\
\midrule

1.0 & 0.5
& 32.47
& 27.71
& 29.90
& 8.30
& 7.31
& 7.77
\\

0.5 & 1.0
& 30.92
& 26.31
& 28.43
& 8.45
& 7.47
& 7.93
\\

2.0 & 2.0
& 32.16
& 28.15
& 30.02
& 8.78
& 7.61
& 8.15
\\

\midrule

\textbf{1.0}
& \textbf{1.0 (Ours)}
& \textbf{33.54}
& \textbf{28.70}
& \textbf{30.93}
& \textbf{9.29}
& \textbf{8.03}
& \textbf{8.62}
\\

\bottomrule
\end{tabular}
\end{table}

\noindent\textbf{Effect of Reward Weights.}
We independently train RGL-RFT with different combinations of $(\alpha,\beta')$, while fixing the format reward weight to $\gamma=0.1$ and keeping all other training and evaluation settings unchanged. As shown in Table~\ref{tab:reward_weights}, equal weighting, i.e., $\alpha=\beta'=1$, achieves the best overall entity and relation performance among the evaluated configurations.
\subsection{Qualitative Results}
\begin{figure*}[t]
    \centering
    \includegraphics[width=1.0\linewidth]{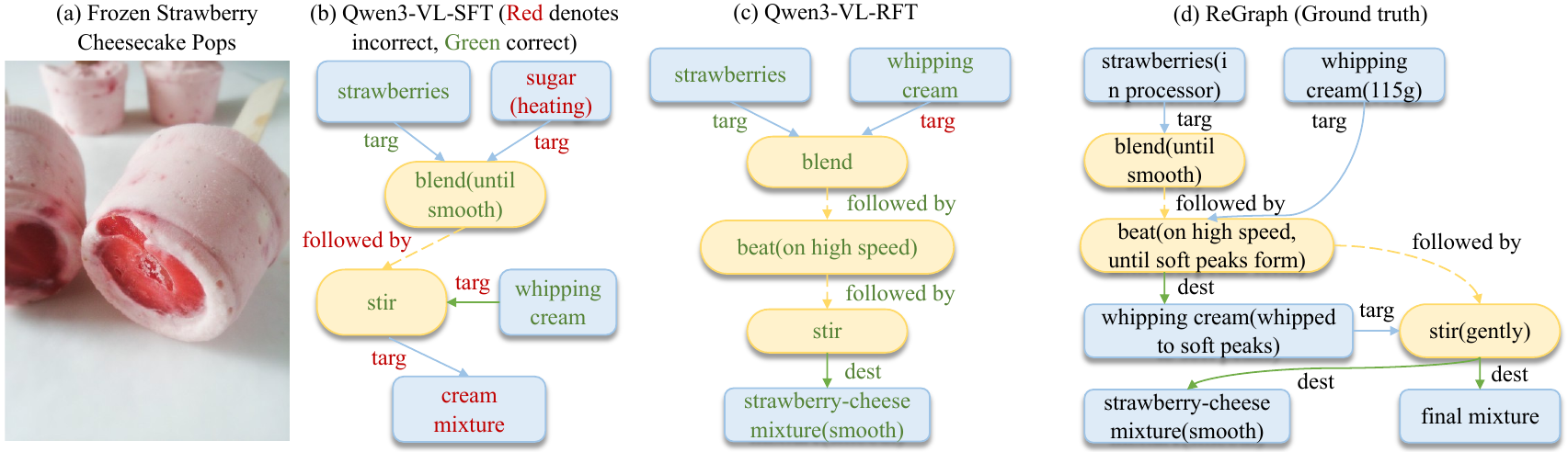}
    \caption{
Qualitative comparison of recipe graph generation.
Qwen3-VL-RFT produces more accurate entities, relations, and procedural dependencies than Qwen3-VL-SFT, resulting in graph structures that more closely match the reference ReGraph. Red denotes incorrect predictions, whereas green denotes correct predictions.
}
    \label{fig:qualitative}
\end{figure*}

Figure~\ref{fig:qualitative} presents a qualitative comparison between the graphs generated by Qwen3-VL-SFT and Qwen3-VL-RFT and the reference ReGraph annotation. Reinforcement fine-tuning clearly improves both the semantic accuracy of graph elements and the procedural consistency of the generated workflow.

As shown in Figure~\ref{fig:qualitative}(b), Qwen3-VL-SFT recovers the main ingredients and the initial \textit{blend} action, but exhibits several characteristic errors: it hallucinates a spurious ingredient, omits the intermediate \textit{beat} step and its resulting whipped state, and produces an incorrect final product (\textit{cream mixture}) through a misassigned \textit{targ} relation. These errors yield a simplified and partially incorrect procedural structure.

In contrast, Figure~\ref{fig:qualitative}(c) shows that Qwen3-VL-RFT produces a graph substantially closer to the reference. It recovers the complete \textit{blend}--\textit{beat}--\textit{stir} action sequence and generates the correct final product with its ingredient state (\textit{strawberry-cheese mixture(smooth)}), yielding \textit{targ}, \textit{dest}, and \textit{followed~by} relations that are consistent with the ground-truth graph.

Although minor discrepancies remain compared with Figure~\ref{fig:qualitative}(d), the overall topology generated by Qwen3-VL-RFT closely matches the annotated procedure. These qualitative results are consistent with the quantitative gains in Table~\ref{tab:main_results}, showing that RGL-RFT improves not only entity and relation prediction but also the generation of more complete, reference-aligned procedural structures.

\begin{figure*}[b]
    \centering

    \includegraphics[width=\textwidth]{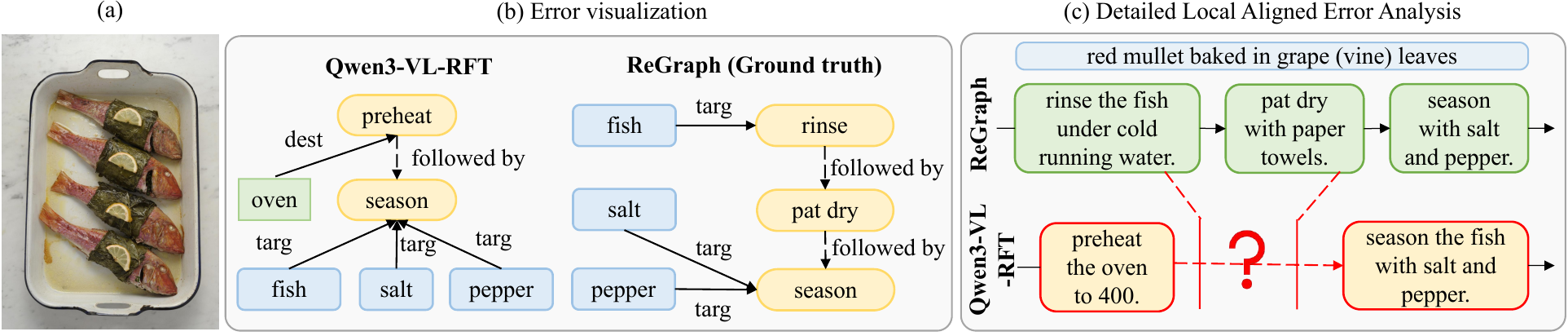}

    \vspace{0.2em}

    \begin{minipage}{0.96\textwidth}
        \small
        \textbf{(1) Visually unobservable procedural information.}
        The model misses invisible preparatory operations (e.g., \textit{rinse} and \textit{pat dry}) while hallucinating a plausible \textit{preheat} action.
    \end{minipage}

    \vspace{0.8em}

    \includegraphics[width=\textwidth]{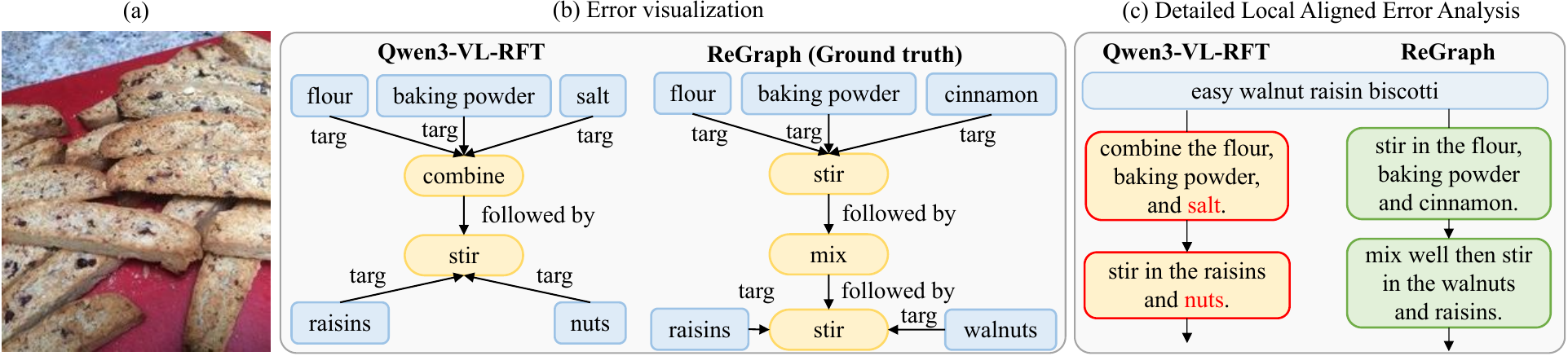}

    \vspace{0.2em}

    \begin{minipage}{0.96\textwidth}
        \small
        \textbf{(2) Commonsense-driven hallucinations and procedural simplification.}
        The model predicts statistically plausible ingredients and simplifies fine-grained procedural structures.
    \end{minipage}

    \caption{
    Representative failure cases of recipe graph generation, illustrating visually unobservable procedural information, commonsense-driven hallucinations, and procedural simplification.
    }
    \label{fig:error_analysis}
\end{figure*}
\subsection{Error Analysis}
\label{sec:error_analysis}
Despite the substantial improvements achieved by RGL-SFT and RGL-RFT, several characteristic failure modes remain. We manually categorize the observed errors into several major types, where one failure case may involve multiple error sources. The most frequently observed failure modes are \textit{visually unobservable procedural information} and \textit{commonsense-driven hallucinations}, followed by procedural simplification and relation-boundary ambiguity.

\textbf{Visually unobservable procedural information.}
This category primarily produces false negatives, reducing recall. Many preparatory operations, such as \textit{rinse}, \textit{soak}, \textit{marinate}, and \textit{pat dry}, leave little or no visual evidence in the final dish. Likewise, ingredients that dissolve or are fully absorbed during cooking (e.g., sugar, butter, or seasoning liquids) become visually indistinguishable after preparation. As illustrated in Case~(1) of Figure~\ref{fig:error_analysis}, the model omits visually imperceptible operations present in the reference such as \textit{rinse} and \textit{pat dry}. The same example also shows a typical commonsense-driven hallucination, where the model predicts a plausible \textit{preheat} operation despite its absence from the reference recipe graph.

\textbf{Commonsense-driven hallucinations.}
Commonsense priors mainly introduce false positives, reducing precision. Rather than matching the annotated reference recipe, the model occasionally predicts ingredients or actions that are statistically common in similar dishes but absent from the reference graph. Case~(2) of Figure~\ref{fig:error_analysis} presents a representative baking example, where the model predicts the common ingredient \textit{salt} instead of the reference ingredient \textit{cinnamon}, reflecting a strong prior learned from baking recipes.

The same example also illustrates entity-granularity ambiguity. The model predicts the generic concept \textit{nuts} instead of the annotated entity \textit{walnuts}. Although visually plausible, this prediction is penalized because the evaluation requires matching the fine-grained reference ingredient specified in the reference graph.

\textbf{Procedural simplification and relation ambiguity.}
Case~(2) further demonstrates a common form of procedural simplification. The reference graph preserves successive \textit{stir--mix--stir} operations inherited from the original recipe, whereas the model compresses them into a shorter \textit{combine--stir} workflow. Although procedurally reasonable, this simplification alters the annotated action boundaries and temporal dependencies, leading to both entity and relation errors.

Overall, the remaining errors are largely attributable to the limited observability of static food images and the gap between visually plausible procedural reasoning and the fine-grained procedural annotations required by ReGraph.


\begin{table}[h]
\centering
\setlength{\abovecaptionskip}{1pt}
\caption{Oracle analysis with annotated recipe context under the deterministic canonical matching protocol.}
\label{tab:upper_bound}
\begin{tabular}{lcccccc}
\hline
\multirow{2}{*}{\textbf{Method}}
& \multicolumn{3}{c}{\textbf{Entity}}
& \multicolumn{3}{c}{\textbf{Relation}}
\\
\cline{2-7}
& \textbf{P} & \textbf{R} & \textbf{F1}
& \textbf{P} & \textbf{R} & \textbf{F1}
\\
\hline

Qwen3-VL-RFT
& 33.54 & 28.70 & 30.93
& 9.29 & 8.03 & 8.62
\\

\quad +GT Ingredients
& 34.65 & 29.99 & 32.15
& 9.67 & 8.43 & 9.01
\\

\quad +GT Instructions
& 55.05 & 35.33 & 43.04
& 20.57 & 14.13 & 16.75
\\

\quad +GT Recipe
& \textbf{63.85} & \textbf{36.18} & \textbf{46.19}
& \textbf{24.09} & \textbf{15.19} & \textbf{18.63}
\\

\hline
\end{tabular}
\end{table}
\subsection{Oracle Context Analysis}
To separate the difficulty of inferring procedural information from images from that of formalizing recipes into graphs, we augment the visual input with increasing levels of annotated context: \textit{+GT Ingredients}, \textit{+GT Instructions}, and the full \textit{+GT Recipe}. As shown in Table~\ref{tab:upper_bound}, adding the ground-truth ingredient list barely helps (Relation F1 8.62$\to$9.01), whereas adding the instructions drives most of the gain (16.75), indicating that the bottleneck lies in recovering the cooking \emph{procedure} rather than merely identifying ingredients. Yet even with the complete recipe, Relation F1 reaches only 18.63\%, showing that formalizing free-form instructions into ReGraph's fine-grained entities, ingredient states, and procedural relations is itself a substantial challenge. The gap between the image-only (8.62\%) and full-context settings should thus not be read as the recoverability of a unique hidden process, but as the combined difficulty of inferring a reference-aligned workflow from limited visual evidence and casting it into an explicit graph. This reinforces the value of ReGraph: procedural knowledge that remains implicit in recipe text is made explicit and measurable through graph formalization.
\section{Limitations}
\label{sec:limitations}
Our study has three main limitations. First, recovering a procedural graph from a single food image is inherently underdetermined: many preparatory operations (e.g., rinsing, marinating, or dissolving ingredients) leave no visual trace in the finished dish, which our error analysis identifies as a major source of missed predictions. The task is therefore framed as generating a plausible, reference-aligned workflow rather than reconstructing a unique hidden process. Second, each image in ReGraph is paired with a single reference graph derived from its Recipe1M recipe, so the reported scores measure alignment with one annotated workflow; a procedurally reasonable graph may still be penalized when it differs from this reference. This single-reference protocol offers a controlled basis for comparison but remains an imperfect proxy for the one-to-many nature of cooking. Third, even with substantial gains, fine-grained ingredient-state capture and relation generation remain the hardest dimensions, and our study is restricted to two 8B backbones. Future work should explore multi-reference or procedural-equivalence evaluation, stronger modeling of visually unobservable steps, and scaling to larger models.
\section{Conclusion}
We present ReGraph, a large-scale recipe graph dataset designed to represent cooking as a structured transformation process. By encoding ingredients, actions, tools, intermediate products, ingredient-state transitions, and procedural dependencies as explicit entities, attributes, and typed relations, ReGraph converts procedural knowledge that is often implicit in free-form recipe language into a compositional graph representation. Building on this representation, we propose Recipe Graph Learning (RGL), a two-stage framework that combines supervised fine-tuning with graph-level reinforcement fine-tuning to generate plausible cooking workflows directly from food images, with RR-CoT traces serving as auxiliary procedural decomposition supervision. Our experiments show that strong text-generation performance does not necessarily imply that a model can explicitly represent how ingredients change state, which actions operate on them, or how cooking steps depend on one another. In contrast, RGL consistently improves reference-aligned entity, ingredient-state, and relation generation across two representative multimodal backbones. ReGraph therefore provides not merely an additional evaluation format, but a process-centric knowledge representation for studying whether multimodal models acquire and express fine-grained procedural knowledge about cooking.

\bibliographystyle{ACM-Reference-Format}
\bibliography{sample-base}

\appendix

\clearpage
\section{Case Study: The ReGraph Dataset}
\label{case_study}
This section presents a complete ReGraph annotation example for the
``cheese stuffed shells'' recipe. The annotation process consists of four
stages: (1) the original recipe input from Recipe1M~\cite{recipe1m},
(2) the initial RR-CoT and recipe graph generated by Claude-Sonnet
4.5~\cite{claude45}, (3) the final entity and relation annotations after
human verification, and (4) a visualization of the verified recipe
graph. The Claude-generated output is retained to illustrate the
LLM-assisted annotation process, while the human-verified annotation is
used as the final ReGraph ground truth.

\subsection{Full Input Data}

\begin{tcolorbox}[
    colback=white,
    colframe=black,
    title={\textbf{Input: Original Recipe}},
    breakable,
]
\ttfamily
\small
Title: cheese stuffed shells.

Ingredients: 2 eggs, beaten, 30 ounces ricotta cheese, lowfat okay, 2 cups shredded mozzarella cheese, lowfat okay, 3/4 cup grated parmesan cheese, lowfat okay, 1/2 cup fresh parsley sprig, chopped, 1 teaspoon salt, 1 dash black pepper, 1/2 ounces shell pasta, jumbo, prepared, 26 ounces spaghetti sauce, lowfat okay, 1/4 cup grated parmesan cheese, lowfat okay.

Instructions: boil shells according to package directions. remove from pot with tongs to avoid tearing and allow to cool on a sheet of wax paper. combine eggs, cheeses, and seasonings in bowl. spoon about 3 tbls. cheese mixture into each cooked shell. spread 2 tbls of the spaghetti sauce in bottom of 13"x9" baking dish. arrange shells in dish over sauce.pour remaining sauce over stuffed shells. sprinkle cheese over sauce. bake in 350f degrees oven for 30 minutes.

\end{tcolorbox}
\subsection{Initial LLM-Assisted Annotation}
Claude-Sonnet 4.5 first analyzes the recipe instructions to extract ordered actions, procedural dependencies, and preliminary entities and relations. These preliminary outputs are subsequently schema-normalized by GPT-4o and manually reviewed and corrected to obtain the final ground truth.

\begin{tcolorbox}[
    colback=gray!5,
    colframe=gray!50,
    title={\textbf{Intermediate Output: Claude-Generated RR-CoT and Recipe Graph}},
    fonttitle=\bfseries,
    breakable,
    left=2mm,
    right=2mm,
    top=1mm,
    bottom=1mm,
    boxrule=0.4pt
]

\begin{lstlisting}[
    basicstyle=\ttfamily\scriptsize,
    breaklines=true,
    breakatwhitespace=false
]
## STEP 1 - ACTION SEQUENCE
a) Numbered Steps with Exact Text:
1. "boil shells according to package directions"
2. "remove from pot with tongs to avoid tearing"
3. "allow to cool on a sheet of wax paper"
4. "combine eggs, cheeses, and seasonings in bowl"
5. "spoon about 3 tbls. cheese mixture into each cooked shell"
6. "spread 2 tbls of the spaghetti sauce in bottom of 13"x9" baking dish"
7. "arrange shells in dish over sauce"
8. "pour remaining sauce over stuffed shells"
9. "sprinkle cheese over sauce"
10. "bake in 350f degrees oven for 30 minutes"
b) Dependency List:
- Step 1 -> Step 2 (must boil shells before removing them)
- Step 2 -> Step 3 (must remove shells before they can cool)
- Step 3 -> Step 5 (needs cooled shells to stuff)
- Step 4 -> Step 5 (needs cheese mixture to stuff shells)
- (Step 3, Step 4) -> Step 5 (stuffing requires both cooled shells and prepared cheese mixture)
- Step 5 -> Step 7 (needs stuffed shells to arrange)
- Step 6 -> Step 7 (needs sauce base in dish before arranging shells over it)
- (Step 5, Step 6) -> Step 7 (arranging requires both stuffed shells and sauce base)
- Step 7 -> Step 8 (must arrange shells before pouring remaining sauce)
- Step 8 -> Step 9 (must pour sauce before sprinkling cheese topping)
- Step 9 -> Step 10 (must complete assembly before baking)
c) Rationale:
- **Step 1 -> Step 2**: Cannot remove shells until they are boiled
- **Step 2 -> Step 3**: Shells must be removed from hot pot before cooling
- **Step 3 -> Step 5**: Shells must be cooled to handle for stuffing
- **Step 4 -> Step 5**: Cheese mixture must be prepared before stuffing shells
- **(Step 3, Step 4) -> Step 5**: Stuffing operation requires both cooled shells (from Step 3) and cheese mixture (from Step 4) to be ready
- **Step 5 -> Step 7**: Must have stuffed shells before arranging them
- **Step 6 -> Step 7**: Sauce base must be spread in dish before arranging shells over it
- **(Step 5, Step 6) -> Step 7**: Arranging requires both stuffed shells (from Step 5) and prepared dish with sauce (from Step 6)
- **Step 7 -> Step 8**: Shells must be arranged before pouring remaining sauce
- **Step 8 -> Step 9**: Sauce must be poured before sprinkling cheese topping
- **Step 9 -> Step 10**: All assembly must be complete before baking
## STEP 2 - KNOWLEDGE GRAPH TRIPLES
a) Action Mapping Table:
- Step 1 -> boil (according to package directions)
- Step 2 -> remove (from pot with tongs to avoid tearing)
- Step 3 -> cool (on a sheet of wax paper)
- Step 4 -> combine
- Step 5 -> spoon (about 3 tbls into each cooked shell)
- Step 6 -> spread (2 tbls in bottom of dish)
- Step 7 -> arrange (in dish over sauce)
- Step 8 -> pour (remaining sauce)
- Step 9 -> sprinkle
- Step 10 -> bake (in 350F degrees oven for 30 minutes)
b) All Entities:
**Ingredients:**
- [I1] eggs (beaten) (source: initial ingredients)
- [I2] ricotta cheese (source: initial ingredients)
- [I3] mozzarella cheese (shredded) (source: initial ingredients)
- [I4] parmesan cheese (grated, 3/4 cup for filling) (source: initial ingredients)
- [I5] parsley (chopped) (source: initial ingredients)
- [I6] salt (source: initial ingredients)
- [I7] pepper (source: initial ingredients)
- [I8] shell pasta (jumbo) (source: initial ingredients)
- [I9] shell pasta (boiled) (source: Step1)
- [I10] shell pasta (cooled) (source: Step3)
- [I11] spaghetti sauce (source: initial ingredients)
- [I12] parmesan cheese (grated, 1/4 cup for topping) (source: initial ingredients)
- [I13] cheese mixture (source: Step4)
- [I14] stuffed shells (source: Step5)
**Tools:**
- [T1] pot (source: Step1, implied)
- [T2] tongs (source: Step2)
- [T3] sheet of wax paper (source: Step3)
- [T4] bowl (source: Step4)
- [T5] baking dish (13"x9") (source: Step6)
- [T6] oven (source: Step10)
**Actions:**
- [A1] boil (according to package directions) (source: Step1)
- [A2] remove (from pot with tongs to avoid tearing) (source: Step2)
- [A3] cool (source: Step3)
- [A4] combine (source: Step4)
- [A5] spoon (about 3 tbls into each cooked shell) (source: Step5)
- [A6] spread (2 tbls in bottom of dish) (source: Step6)
- [A7] arrange (in dish over sauce) (source: Step7)
- [A8] pour (remaining sauce) (source: Step8)
- [A9] sprinkle (source: Step9)
- [A10] bake (in 350F degrees oven for 30 minutes) (source: Step10)
c) All Relationships:
**targ relations:**
- [R1] (I8, targ, A1) - shell pasta -> boil (source: Step1)
- [R2] (I9, targ, A2) - shell pasta (boiled) -> remove (source: Step2)
- [R3] (I9, targ, A3) - shell pasta (boiled) -> cool (source: Step3)
- [R4] (I1, targ, A4) - eggs (beaten) -> combine (source: Step4)
- [R5] (I2, targ, A4) - ricotta cheese -> combine (source: Step4)
- [R6] (I3, targ, A4) - mozzarella cheese (shredded) -> combine (source: Step4)
- [R7] (I4, targ, A4) - parmesan cheese (grated, 3/4 cup) -> combine (source: Step4)
- [R8] (I5, targ, A4) - parsley (chopped) -> combine (source: Step4)
- [R9] (I6, targ, A4) - salt -> combine (source: Step4)
- [R10] (I7, targ, A4) - pepper -> combine (source: Step4)
- [R11] (I13, targ, A5) - cheese mixture -> spoon (source: Step5)
- [R12] (I10, targ, A5) - shell pasta (cooled) -> spoon (source: Step5)
- [R13] (I11, targ, A6) - spaghetti sauce -> spread (source: Step6)
- [R14] (I14, targ, A7) - stuffed shells -> arrange (source: Step7)
- [R15] (I11, targ, A8) - spaghetti sauce -> pour (source: Step8)
- [R16] (I12, targ, A9) - parmesan cheese (grated, 1/4 cup) -> sprinkle (source: Step9)
- [R17] (T5, targ, A10) - baking dish -> bake (source: Step10)
**dest relations:**
- [R18] (A1, dest, I9) - boil -> shell pasta (boiled) (source: Step1)
- [R19] (A2, dest (onto), T3) - remove -> sheet of wax paper (source: Step2)
- [R20] (A3, dest, I10) - cool -> shell pasta (cooled) (source: Step3)
- [R21] (A4, dest (in), T4) - combine -> bowl (source: Step4)
- [R22] (A4, dest, I13) - combine -> cheese mixture (source: Step4)
- [R23] (A5, dest, I14) - spoon -> stuffed shells (source: Step5)
- [R24] (A6, dest (in bottom of), T5) - spread -> baking dish (source: Step6)
- [R25] (A7, dest (in dish over sauce), T5) - arrange -> baking dish (source: Step7)
- [R26] (A8, dest (over stuffed shells in), T5) - pour -> baking dish (source: Step8)
- [R27] (A9, dest (over sauce in), T5) - sprinkle -> baking dish (source: Step9)
- [R28] (A10, dest (into), T6) - bake -> oven (source: Step10)
**followed by relations:**
- [R29] (A1, followed by, A2) - boil -> remove (source: Step1->Step2)
- [R30] (A2, followed by, A3) - remove -> cool (source: Step2->Step3)
- [R31] (A3, followed by, A5) - cool -> spoon (source: Step3->Step5, cooled shells needed for stuffing)
- [R32] (A4, followed by, A5) - combine -> spoon (source: Step4->Step5, cheese mixture needed for stuffing)
- [R33] (A5, followed by, A6) - spoon -> spread (source: Step5->Step6, logical sequence to prepare dish)
- [R34] (A6, followed by, A7) - spread -> arrange (source: Step6->Step7, sauce base needed before arranging)
- [R35] (A7, followed by, A8) - arrange -> pour (source: Step7->Step8)
- [R36] (A8, followed by, A9) - pour -> sprinkle (source: Step8->Step9)
- [R37] (A9, followed by, A10) - sprinkle -> bake (source: Step9->Step10)
\end{lstlisting}

\end{tcolorbox}
\subsection{Human-Verified ReGraph Annotation}
The initial LLM-generated annotation is manually verified before being used as ground truth. Three types of corrections are applied. First, the \texttt{source\_step} of several ingredient entities is realigned from the generic ``initial ingredients'' listing to the step in which each ingredient is actually consumed (e.g., the filling ingredients I1--I7 are reassigned to Step~4, and the topping cheese I12 to Step~9), improving procedural traceability. Second, a missing final-state entity is completed: the initial annotation leaves the \texttt{bake} action without an output entity, so we add the terminal ingredient I15 (\emph{stuffed shells}, baked) together with its \texttt{dest} relation, making the ingredient-state chain end-to-end complete. Third, unattached tool usages are connected: the tool interactions for \texttt{boil} (pot) and \texttt{remove} (tongs) are added so that no tool entity is left isolated. The verified annotation is shown below, with the newly added entity and relations highlighted.
\begin{tcolorbox}[
    colback=blue!2,
    colframe=blue!45!black,
    title={\textbf{Final Output: Human-Verified Entities and Relations}},
    fonttitle=\bfseries,
    breakable,
    left=2mm, right=2mm, top=1mm, bottom=1mm,
    boxrule=0.5pt
]
\begin{Verbatim}[
    fontsize=\scriptsize,
    breaklines=true,
    breakanywhere=true,
    commandchars=\\\{\}
]
ALL ENTITIES
Ingredients:
[I1]  eggs (attribute: beaten)
[I2]  ricotta cheese
[I3]  mozzarella cheese (attribute: shredded)
[I4]  parmesan cheese (attribute: grated, 3/4 cup for filling)
[I5]  parsley (attribute: chopped)
[I6]  salt
[I7]  pepper
[I8]  shell pasta (attribute: jumbo)
[I9]  shell pasta (attribute: boiled)
[I10] shell pasta (attribute: cooled)
[I11] spaghetti sauce
[I12] parmesan cheese (attribute: grated, 1/4 cup for topping)
[I13] cheese mixture
[I14] stuffed shells
\textcolor{red}{[I15] stuffed shells (attribute: baked)  <-- added}
Tools:
[T1] pot
[T2] tongs
[T3] sheet of wax paper
[T4] bowl
[T5] baking dish (attribute: 13"x9")
[T6] oven
Actions:
[A1]  boil (attribute: according to package directions)
[A2]  remove (attribute: from pot with tongs to avoid tearing)
[A3]  cool (attribute: on a sheet of wax paper)
[A4]  combine
[A5]  spoon (attribute: about 3 tbls into each cooked shell)
[A6]  spread (attribute: 2 tbls in bottom of dish)
[A7]  arrange (attribute: in dish over sauce)
[A8]  pour (attribute: remaining sauce)
[A9]  sprinkle
[A10] bake (attribute: in 350F degrees oven for 30 minutes)
ALL RELATIONS
targ:
[R1]  (I8,  targ, A1)     [R2]  (I9,  targ, A2)
[R3]  (I9,  targ, A3)     [R4]  (I1,  targ, A4)
[R5]  (I2,  targ, A4)     [R6]  (I3,  targ, A4)
[R7]  (I4,  targ, A4)     [R8]  (I5,  targ, A4)
[R9]  (I6,  targ, A4)     [R10] (I7,  targ, A4)
[R11] (I13, targ, A5)     [R12] (I10, targ, A5)
[R13] (I11, targ, A6)     [R14] (I14, targ, A7)
[R15] (I11, targ, A8)     [R16] (I12, targ, A9)
[R17] (T5,  targ, A10)
\textcolor{red}{[R18] (T2,  targ, A2)  <-- added}
dest:
[R19] (A1,  dest, I9)
\textcolor{red}{[R20] (A1,  dest, T1) [relation attribute: in]  <-- added}
[R21] (A2,  dest, T3) [relation attribute: onto]
[R22] (A3,  dest, I10)
[R23] (A4,  dest, T4) [relation attribute: in]
[R24] (A4,  dest, I13)
[R25] (A5,  dest, I14)
[R26] (A6,  dest, T5) [relation attribute: in bottom of]
[R27] (A7,  dest, T5) [relation attribute: in dish over sauce]
[R28] (A8,  dest, T5) [relation attribute: over stuffed shells in]
[R29] (A9,  dest, T5) [relation attribute: over sauce in]
[R30] (A10, dest, T6) [relation attribute: into]
\textcolor{red}{[R31] (A10, dest, I15)  <-- added}
followed by:
[R32] (A1, followed by, A2)     [R33] (A2, followed by, A3)
[R34] (A3, followed by, A5)     [R35] (A4, followed by, A5)
[R36] (A5, followed by, A6)     [R37] (A6, followed by, A7)
[R38] (A7, followed by, A8)     [R39] (A8, followed by, A9)
[R40] (A9, followed by, A10)
\end{Verbatim}
\end{tcolorbox}
\begin{figure*}[!t]
    \centering
    \includegraphics[width=0.8\textwidth]{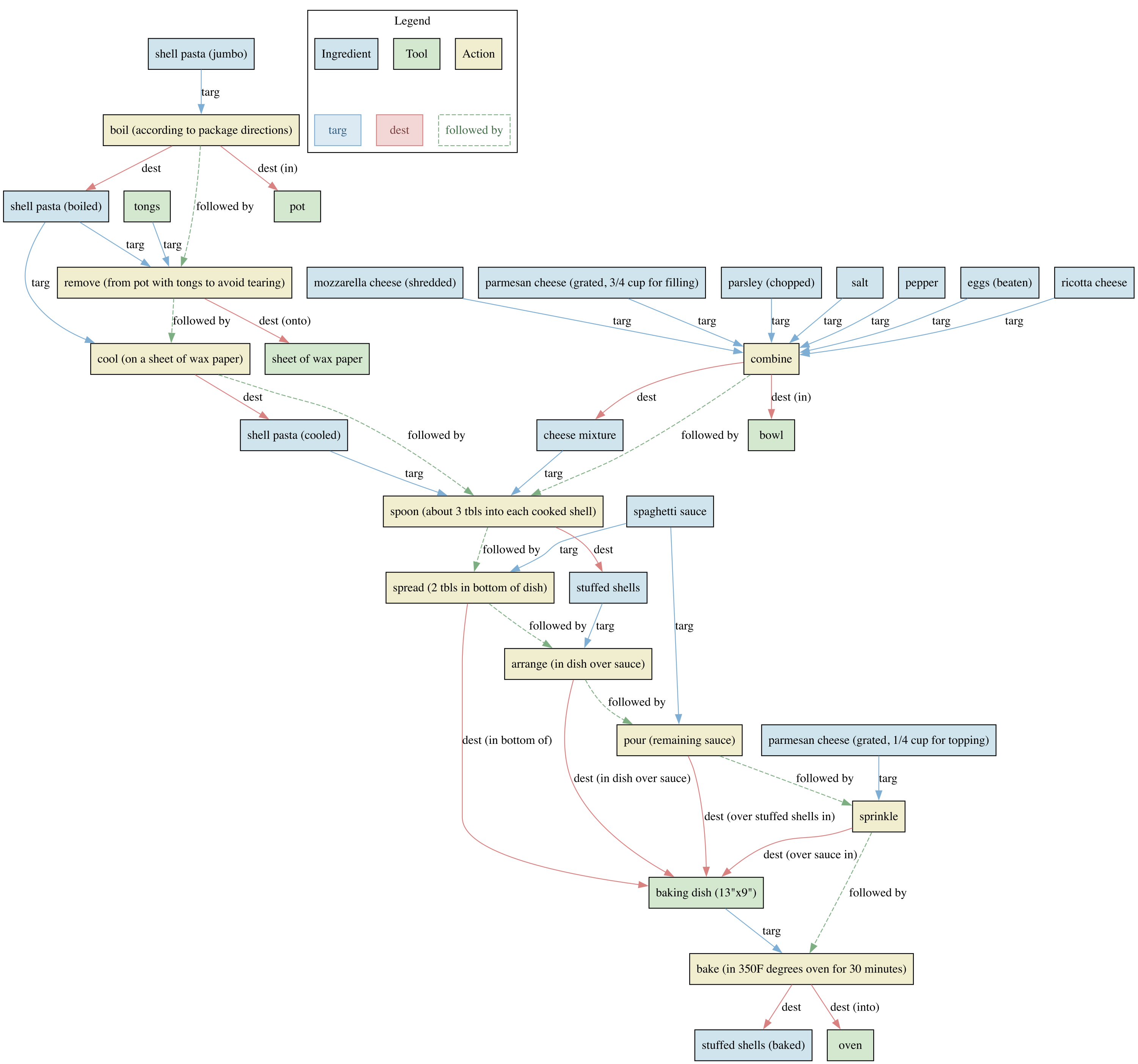}
    \caption{
        Human-verified ReGraph annotation for the ``cheese stuffed shells'' recipe. The graph explicitly represents ingredient-state
        transformations and procedural ordering throughout the cooking
        workflow.
    }
    \label{fig:kg_case_study}
\end{figure*}
\subsection{Verified Recipe Graph Visualization}

Figure~\ref{fig:kg_case_study} visualizes the final human-verified
annotation. Ingredient and intermediate-state entities are shown in
blue, action entities in yellow, and tool entities in green.
The \texttt{targ} relation connects an input entity to the action applied
to it, \texttt{dest} connects an action to its destination or resulting
state, and \texttt{followed by} represents procedural ordering between
successive cooking actions.

\section{Prompt Overview for ReGraph Construction}
\label{prompt}

ReGraph is constructed through a two-stage prompting pipeline. Claude-Sonnet 4.5 \cite{claude45} first performs procedural reasoning and extracts preliminary entities and relations from recipe text. GPT-4o \cite{gpt4o} then performs schema normalization only, converting these preliminary outputs into a unified graph schema. Rather than reproducing the complete prompts, we summarize the key objectives below.

\begin{tcolorbox}[
title=Stage 1: Procedural Reasoning and Graph Construction,
colback=gray!5,
colframe=black,
boxrule=0.5pt]

\textbf{Input:}
Recipe title, ingredient list, and cooking instructions.

\medskip

\textbf{Objectives:}
\begin{itemize}[nosep,leftmargin=1.3em]
\item Generate an RR-CoT by decomposing the recipe into executable steps and identifying temporal/procedural dependencies.
\item Construct a recipe graph by extracting \texttt{ingredient}, \texttt{tool}, and \texttt{action} entities together with \texttt{targ}, \texttt{dest}, and \texttt{followed by} relations.
\item Explicitly model ingredient state changes, preserve procedural order, and record source-step traceability.
\end{itemize}

\textbf{Output:}
RR-CoT, raw entity list, and raw relation list.

\end{tcolorbox}

\vspace{-0.5em}

\begin{tcolorbox}[
title=Stage 2: Schema Normalization,
colback=gray!5,
colframe=black,
boxrule=0.5pt]

\textbf{Input:}
RR-CoT, preliminary entities, and preliminary relations.

\medskip

\textbf{Objectives:}
\begin{itemize}[nosep,leftmargin=1.3em]
\item Convert the preliminary outputs into the predefined graph schema.
\item Ensure consistent field formatting and validate identifiers.
\item Preserve source-step traceability and export Graphviz and JSON annotations.
\end{itemize}

\textbf{Output:}
Graphviz graph, entity JSON, and relation JSON.

\end{tcolorbox}

\noindent\textbf{Representative Annotation Schema}

\begin{lstlisting}[basicstyle=\ttfamily\footnotesize]
Entity:
{
  entity_id,
  entity_type,
  entity_name,
  attributes,
  source_step
}

Relation:
{
  head_entity_id,
  relation_type,
  tail_entity_id,
  relation_attributes,
  source_step
}
\end{lstlisting}
\end{document}